\documentclass[sigconf]{acmart}
\acmSubmissionID{}

\usepackage[utf8]{inputenc}
\usepackage{url}
\usepackage{subfig}
\usepackage{graphicx}
\graphicspath{{./figures/},{./figures/plots/}}
\usepackage{rotating}
\usepackage{colortbl}
\usepackage{multirow}
\usepackage{diagbox}
\usepackage{tablefootnote}
\usepackage[export]{adjustbox}

\usepackage{dcolumn}
\newcolumntype{d}[1]{D{.}{.}{#1}}

\usepackage{bbm, bbold}

\usepackage[bottom]{footmisc}

\usepackage{algorithm}
\usepackage{algorithmicx}
\usepackage{algpseudocode}
\usepackage[ruled,vlined,linesnumbered,algo2e]{algorithm2e}

\usepackage{enumitem}
\setlist[itemize]{leftmargin=*}

\newcommand\REVISION[1]{#1}
\newcommand\ADD[1]{\textcolor{black}{#1}}
\newcommand\Yue[1]{\textcolor{black}{#1}}

\newcommand{\loss}{\mathcal{L}}
\newcommand{\image}{\mathcal{I}}
\newcommand{\encoder}{\mathcal{E}}
\newcommand{\decoder}{\mathcal{D}}
\newcommand{\normal}{\mathcal{N}}

\begin{document}

\title[EyeFormer]{EyeFormer: Predicting Personalized Scanpaths with Transformer-Guided Reinforcement Learning}

\author{Yue Jiang}
\email{yue.jiang@aalto.fi}
\authornote{Contributed equally.}
\affiliation{%
  \institution{Aalto University}
  \country{Finland}
}

\author{Zixin Guo}
\email{zixin.guo@aalto.fi}
\authornotemark[1]
\affiliation{%
  \institution{Aalto University}
  \country{Finland}
}

\author{Hamed R. Tavakoli}
\email{hamed.rezazadegan_tavakoli@nokia.com}
\affiliation{%
  \institution{Nokia Technologies}
  \country{Finland}
}

\author{Luis A. Leiva}
\email{name.surname@uni.lu} %
\affiliation{%
  \institution{University of Luxembourg}
  \country{Luxembourg}
}

\author{Antti Oulasvirta}
\email{antti.oulasvirta@aalto.fi}
\affiliation{%
  \institution{Aalto University}
  \country{Finland}
}

\begin{abstract}
From a visual perception perspective, modern graphical user interfaces (GUIs) comprise a complex graphics-rich two-dimensional visuospatial arrangement of text, images, and interactive objects such as buttons and menus.
While existing models can accurately predict regions and objects that are likely to attract attention ``on average'', 
so far there is no scanpath model capable of predicting scanpaths for an individual.
To close this gap, we introduce EyeFormer, which leverages a Transformer architecture as a policy network 
to guide a deep reinforcement learning algorithm that controls gaze locations.
Our model has the unique capability of producing personalized predictions when given a few user scanpath samples.
It can predict full scanpath information, including fixation positions and duration, 
across individuals and various stimulus types.
Additionally, we demonstrate applications in GUI layout optimization driven by our model. 
Our software and models will be publicly available.

\end{abstract}

\begin{teaserfigure}
  \def\w{\linewidth}
  \centering
 \includegraphics[width=\w]{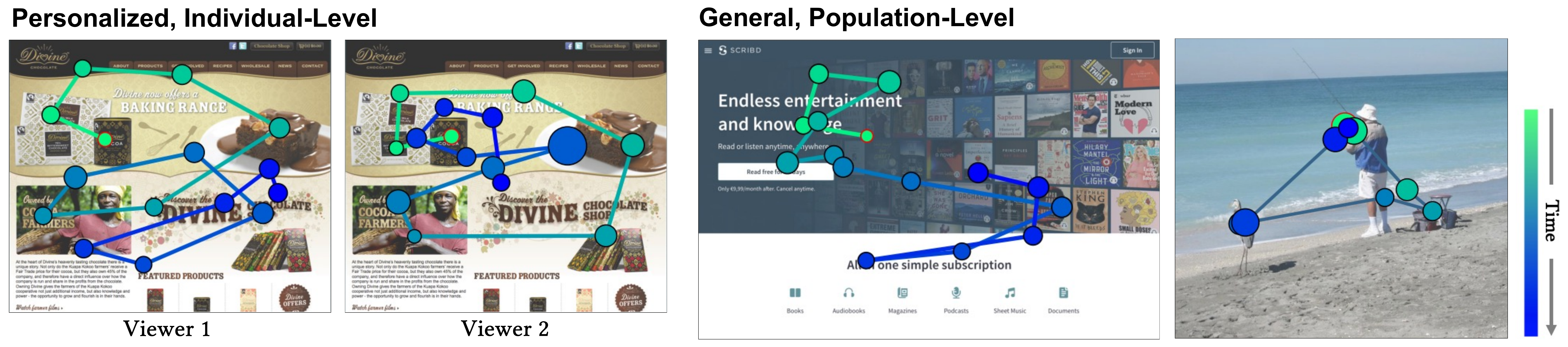}
 \vspace{-6mm}
\caption{\ADD{
We present a novel predictive model of scanpaths that, for the first time, accounts for individual differences and diverse stimuli, 
including both natural scenes (e.g., landscapes, buildings) and artificial scenes (e.g., user interfaces, information graphics). 
Predictions cover the spatial and temporal characteristics of a scanpath; that is, a sequence of fixation locations together with their duration.  
The model can generate ``an average'' scanpath reflecting population-level tendencies
%creating a ``general'' scanpath for each input image, outperforming prior models.
%The model can generate 
as well as scanpaths personalized for individual viewers given a few scanpath samples, 
reflecting each viewer's unique preferences and viewing behaviors. 
%For natural scenes, we predict viewing time up to 3 seconds~\cite{xu2014predicting}, and for GUIs until 7 seconds~\cite{jiang2023ueyes}.
%EyeFormer is the first to generate scanpaths for visually very different stimuli (i.e. both natural and artificial scenes including UIs) and for both population-level and individual-level prediction.
These illustrative plots use a color gradient, from green to blue, to denote the temporal progression of each scanpath. 
Fixation points are denoted by circles, the radii of which represent fixation duration.
}
\Description{Sample results of predicted personalized, individual-level scanpaths and general, population-level scanpaths.}
} 
  \label{fig:teaser}
\end{teaserfigure}

\maketitle

% \begin{figure}[b]
% \noindent\fbox{%
% \parbox{\dimexpr\linewidth-2\fboxsep-2\fboxrule\relax}{%
% \begin{tabular}{l}
% The word count of this paper is \textcolor{blue}{8931}.
% \end{tabular}
% }%
% }
% \end{figure}

\section{Introduction}

A fundamental goal in the design of graphical user interfaces (GUIs) is to guide users' attention toward discovering relevant information and possibilities for interaction~\cite{Rosenholtz11}. 
However, modern GUIs' graphics-rich visuospatial arrangement of text, images, animations, and numerous interactive objects, such as buttons and menus~\cite{jiang2023future, jiang2022computational, jiang2024computational2, jiang2024computational}, makes it increasingly difficult to predict and direct visual attention for different people. 
Furthermore, GUI design is not the only factor in eye movements; 
idiosyncratic factors such as expectations and user-specific attention strategies exert an influence. 
Therefore, predicting how the attention of a given user is going to evolve over time is technically challenging. 
Breakthroughs in this space would afford the design of personalized visual flows, reduce clutter, and make user interfaces more engaging, usable, and visually appealing overall~\cite{Still10}.

Large individual differences have been reported in viewing patterns~\cite{jiang2023ueyes}. To avoid the fallacy of predicting an ``average scanpath'' with no correspondence among actual viewers' behavior, models should capture individual variability in viewing patterns. Such models would open up new applications: they could be used to make predictions for an interesting viewer segment or even for individual viewers.
Effective solutions to this challenge would advance applications of human attention models in visual computing and related domains.

Prior work has primarily focused on \textit{saliency maps}, which represent eye movement data as density maps over images~\cite{itti1998model}. 
However, as static representations, they overlook temporal information. 
In contrast, \textit{scanpaths} contain a wealth of information on fixations, retaining information about the order in which objects and regions are attended, as well as their respective duration~\cite{Mondal_2023_CVPR, jiang2023ueyes, chen2021predicting}. Scanpaths are, therefore, first-order models of human vision from which second-order measurements such as saliency maps can be derived, but not the other way around.
In addition, prior research on scanpath modeling has predominantly centered on natural scenes.
To make these models more generalizable, it is critical to develop unified models that can work with different kinds of stimuli. 
The challenge is that visual attention is highly adaptive to the type of stimuli, and viewing patterns can therefore be widely different on, e.g., websites and mobile GUIs~\cite{leiva2020understanding}.
Any improvement in scanpath predictive modeling will immediately carry over to practical applications. 
For example, such models will allow designers to understand visual flows 
and adjust their designs to encourage users to view the GUI elements in the desired order~\cite{pang2016directing}.

% research gap
We present \emph{EyeFormer}, a scanpath model for free-viewing tasks, 
which can accurately predict both population-level and individual-level spatiotemporal characteristics of viewing behaviors 
across different types of stimuli. 
We formulate the placement of fixations as a reinforcement learning (RL) problem and use a Transformer architecture as a policy network guiding the selection of the subsequent fixation. Transformers have proven effective in various tasks, ranging from language to vision~\cite{Wang2019, ViT2021}. 
Critically, their capability of modeling long sequences~\cite{Tieyuan2023} makes them suitable for scanpath prediction.
EyeFormer's Transformer-guided deep RL approach was designed to address three critical shortcomings of existing approaches.
%Existing models of scanpath prediction fall short in predicting scanpaths in various respects. 
First, predicting the order of fixations from saliency maps, utilizing their probability distribution~\cite{martin2022probabilistic, de2022scanpathnet}, is inherently hard because temporal information is not included in such representations. 
A second issue stems from post-processing steps introduced to prevent the clustering of fixations in density-based approaches~\cite{jiang2023ueyes}, such as inhibition of return~\cite{itti1998model}. This method prevents repeated fixation at a previously identified position within a saliency map. 
However, because these steps are not learnable from data~\cite{Emami24_etra}, one cannot formulate proper loss terms derived from them. 
Third, although recent advances such as PathGAN~\cite{Assens2018pathgan} have brought progress toward handling fixation duration, accuracy in predicting fixation points remains limited because these techniques often generate points outside the areas of interest. 

EyeFormer is the first model to predict full scanpaths at \emph{both} individual and population levels, 
including fixations with coordinates and durations.
It has the unique capability of predicting an individual's viewing behaviors when given a few sample scanpaths from such an individual.
Moreover, we show that EyeFormer compares favorably against prior scanpath models on the vast majority of metrics for both GUIs and natural scenes on population-level scanpath prediction. 
EyeFormer accurately predicts both spatial (where) and temporal (order, duration) characteristics of scanpaths on both GUIs and natural scenes. 
Further, we demonstrate an application of personalized scanpath prediction in creating personalized GUI layouts, which considers both the viewing order and fixation density of GUI elements. In addition, we can generate a single optimized GUI layout that shows minimal variability across individuals to attract attention to desired elements.
In summary, this paper makes the following contributions:
\begin{enumerate}
\item We propose EyeFormer, a deep RL approach incorporating the Transformer architecture 
that predicts both spatial and temporal characteristics of scanpaths, 
thus yielding a comprehensive understanding of viewers' viewing behaviors.
\item Our model can generate personalized scanpaths using only a few scanpath samples from the relevant viewer, 
whereby the model can capture and reflect each viewer's viewing behaviors and preferences. 
\item We offer quantitative and qualitative evaluations demonstrating that the proposed model compares favorably against state-of-the-art scanpath models on GUIs and natural scenes for population-level scanpath prediction.
\item We demonstrate an application of personalized UI optimization facilitated by personalized scanpath prediction. 
\end{enumerate}

\section{Related Work}

Scanpath models predict sequences of fixations for a given image.
This task is more challenging than predicting (dense) saliency maps because the order of the (discrete) fixations must be retained. 
All previous research has concentrated on modeling scanpath patterns at the population level, i.e., an average user model, and no prior work has focused on the prediction of personalized scanpaths. Thus, we conduct a comprehensive review of existing approaches to population-level scanpath prediction, aiming to extend these techniques to the individual level using a novel Transformer-based architecture.
Prior work can be distinguished into three main approaches, 
based on how they have attempted to derive sequential information: 
(i)~computing it \emph{post-hoc} from densities captured in saliency maps; 
(ii)~directly predicting sequences; 
and (iii)~formulating this as a sequential control problem via RL.
For evaluation, we will compare our EyeFormer model against a number of models mentioned next.

\subsection{Saliency Map-based Scanpath Prediction}

Saliency maps, although they do not explicitly contain temporal information, can be used to estimate scanpaths. \citet{itti1998model} introduced an Inhibition of Return (IOR) mechanism to this end. It samples a starting fixation and ``discourages'' future fixations from returning to it, thus producing a sequence. This idea has been evolved in several papers~\cite{REZAZADEGANTAVAKOLI2013686, wloka2018active, Chen2018_IJCAI, Xia2019, assens2017saltinet, kummerer2022deepgaze, martin2022probabilistic, wang2023scanpath}. However, these methods face challenges in three aspects: they (i)~neglect some key temporal aspects, specifically fixation duration, (ii)~lack a coherent ranking order of the fixations, and (iii)~cannot be used in loss terms due to being non-differentiable.

\subsection{Predicting Fixation Sequences}

Some papers have attempted to solve these challenges by sequentially sampling fixations from pre-generated Gaussian distributions 
and by integrating well-designed supervised loss terms. 
This strategic choice enforces a meaningful order among the fixation points. 
For instance, IOR-ROI~\cite{sun2019visual, Chen2018_IJCAI}, ScanpathNet~\cite{de2022scanpathnet}, and Visual ScanPath Transformer~\cite{qiu2023visual} predict fixation distributions by a parameterized Gaussian mixture for generating fixation distributions. 
Gazeformer~\cite{Mondal_2023_CVPR} presents a Transformer-based architecture for goal-oriented viewing tasks.
ScanDMM~\cite{Sui_2023_CVPR} utilizes a Markov model to represent fixation distributions. These models suffer from accumulated errors~\cite{ranzato2015sequence}, where the errors in previously generated points affect the prediction of the following points.
Other models directly predict fixation sequences. 
\citet{Verma2019} predicted fixations using a grid-based representation in which each fixation point connects to a specific region. 
PathGAN~\cite{Assens2018pathgan} and ScanGAN~\cite{martin2022scangan360} applied a GAN-based architecture to generate fixation sequences. 
GAN-based scanpath models have limitations, including fixation points clustering towards the image center, 
and reduced accuracy in predicting fixation points (sometimes outside areas of interest).
NeVA~\cite{schwinn2022behind} addressed downstream visual tasks with unseen
datasets by relying on existing pretrained models for the task rather than simulating human scanpaths.

\subsection{Reinforcement Learning for Scanpath Prediction}

%To avoid accumulated errors in scanpath prediction, 
RL has been studied as a means to formulate scanpaths as a sequential control problem~\cite{mousavi2017learning, edc41706ff574381b7b93fe1403cd4ec}. 
For example, \citet{minut2001reinforcement} 
proposed an RL model for visual search tasks wherein an agent learns 
to focus on relevant areas to locate a target object in a cluttered environment.
\citet{ognibene2008reinforcement}
employed RL using an eye-centered potential action map that accumulates potential target locations over fixations.
\citet{yang2020predicting} used inverse RL for predicting the scanpaths involved in a visual search task, and
\citet{xu2018predicting} applied deep RL specifically to predict head-movement-related scanpaths for panoramic videos.

\Yue{
Recent work \citet{chen2021predicting} discretized fixation positions by dividing the images into grids and predicting the grid corresponding to the fixation.
Inspired by the policy gradient in discrete token generation within visual captioning~\cite{rennie2017self}, 
\citet{chen2021predicting} adopted the policy gradient to optimize for non-differentiable metrics in their discrete token generation.
The position discretization offers the advantage of optimizing a finite set of discrete actions, instead of an infinite continuous space.
However, the artificiality of discretization brings a coarser fixation representation, leading to precision and information loss. Continuous control is challenging since a continuous range of control contains an infinite number of feasible actions~\cite{tang2020discretizing}.
We construct the fixation prediction as a continuous value generation task and resort to parametric functions for Gaussian distributions over actions, optimized by our designed rewards. }

\begin{figure}[t]
  \def\w{\linewidth}
  \centering

 \includegraphics[width=\w]{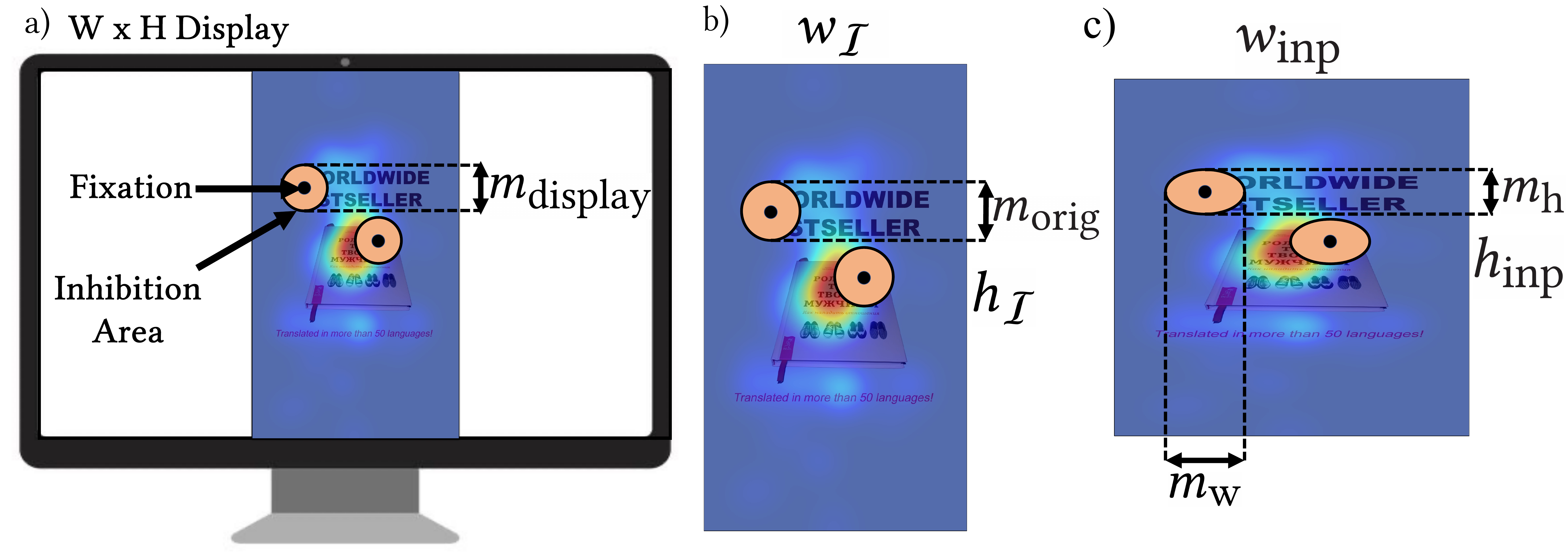}
\caption{The mechanism of adapting the inhibition-of-return area in a display to compute the salient-value reward involves modifying the radius of the inhibition area, which is determined by the disparity between the size of the saliency map and the image size on the display. a)~The diameter of the display's inhibition areas $m_{\textrm{display}}$ is commensurate with human's visual angle. 
b)~We then compute $m_{\textrm{orig}}$, the diameter for the corresponding inhibition areas for the input image with size $w_\image \times h_\image$.
c)~The image needs to be resized to dimensions $w_\textrm{inp} \times h_\textrm{inp}$, which corresponds to the input image size required by the model. Thus, the inhibition areas are rendered, accordingly, as ellipses with radii $m_w$ and $m_h$.}
  \Description{We show how we compute and adapt the inhibition-of-return mechanism from the display to t.}
  \label{fig:ior}
\end{figure}

\section{Method}

\begin{figure*}[t]
  \def\w{\linewidth}
  \centering

 \includegraphics[width=\w]{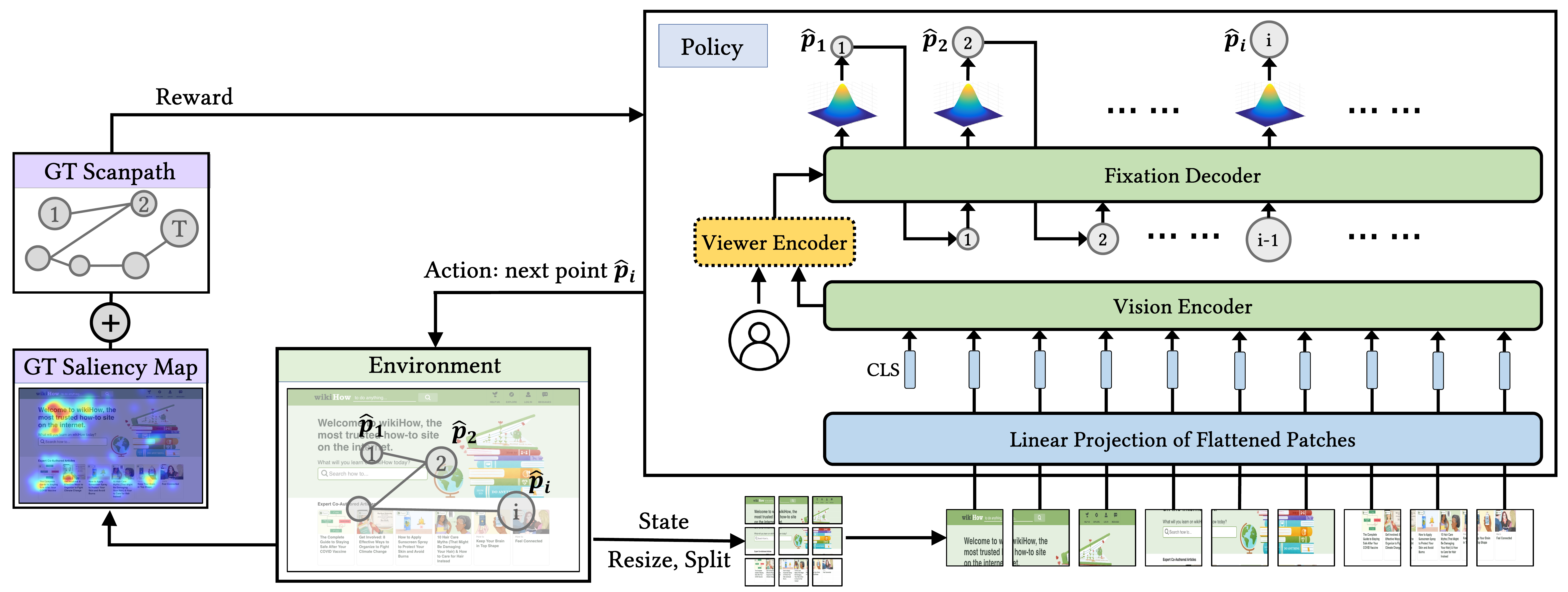}
\caption{
Overview of our Transformer-guided Reinforcement Learning framework for scanpath prediction. It comprises several components: the environment, which produces the state of the input image and previous fixation points; the Transformer model, which furnishes the policy; the policy-generated action, predicting the next point in the scanpath; and the reward function (obtained from evaluating the action against ground truth), through which the policy gets updated. 
Within the Transformer policy model, the image patches, resized and split from the input image, are fed to the vision encoder to get the image embedding; the viewer encoder generates the viewer embedding to distinguish between viewers (only for individual-level prediction); the fixation decoder takes the image and viewer embeddings along with previously generated fixations to sequentially generates the following points along the scanpath.
During training, the model begins with sampling the next point from the distribution generated by the policy in light of the current \textit{state}. Then, this sampled point is used to update the state of the environment, and incorporating the \textit{reward} indicated via ground truth serves to update the Transformer policy model. During testing, we directly use the policy model to generate the scanpaths.
}
\Description{The overview of our Transformer-guided Reinforcement Learning framework for scanpath prediction. It comprises several components: the environment, which produces the state of the input image and previous fixation points; the Transformer model, which furnishes the policy; the policy-generated action, predicting the next point in the scanpath; and the reward function (obtained from evaluating the action against ground truth), through which the policy gets updated. Within the Transformer policy model, the image patches, resized and split from the input image, are fed to the vision encoder to get the image embedding; the viewer encoder generates the viewer embedding to distinguish between viewers (only for individual-level prediction); the fixation decoder takes the image and viewer embeddings along with previously generated fixations to sequentially generates the following points along the scanpath.}
  \label{fig:rl}
\end{figure*}

EyeFormer performs scanpath predictions by sequentially generating fixation points taking both preceding fixations and the scene into account as its state. 
The main challenges we address are related to 1)~generating both spatial and temporal information on fixation points with parametric distributions; 2)~optimizing a scanpath with non-differentiable objectives; and 3)~capturing individual viewing differences to predict personalized scanpaths. 
To address these challenges, we propose a Transformer-guided RL approach (\autoref{fig:rl}) with two key ideas: i)~We use the Transformer architecture to capture long-range sequential dependencies from previous fixations with Gaussian distributions~\cite{vaswani2017attention};
ii)~We use RL instead of directly optimizing the loss since some loss terms are not differentiable, and thus cannot be directly optimized. The RL framework enables us to optimize scanpaths with non-differentiable reward functions~\cite{sutton2018reinforcement}, such as the computation of salient values with inhibition of return (IOR).

\subsection{Problem Formulation}

Given an image $\image$, we generate a scanpath of length $T$, 
i.e., a sequence of ordered fixation points $p_{1:T}=(p_1, \dots, p_T)$ 
capturing the spatial and temporal information of the human gaze. 
%\ADD{~\autoref{fig:scanpath_example} illustrates generating the scanpath from the starting point to the $T$th point.}
Each fixation $p_i = (x_i, y_i, t_i)$ is a three-dimensional vector representing the normalized point coordinates $x_i\in [0,1]$ and $y_i \in [0,1]$ alongside the third dimension, fixation duration expressed as $t_i \in (0, +\infty)$.

\subsection{Environment, State, and Action}

Our predictive model acts as an \textit{agent} that interacts with the \textit{environment}, 
where the latter produces the \textit{state} of both input image $\image$ and previous fixation points.
The $\theta$ parameters dictate the \textit{policy}, $\pi_\theta$, 
whereby the model generates an \textit{action} as a prediction for the next fixation point $\hat{p}_{i}$, 
sampled from the distribution produced by the policy model.
This process is formulated as $\pi_{\theta}(\hat{p}_{i}|\hat{p}_{1:i-1},\image)$.

\subsection{Reward Function}

After each action, the agent receives a salient-value reward $r_\mathrm{sal}$ that expresses the action's contribution to the full scanpath.
 Once the entire scanpath is generated, the agent is exposed to a reward $r_\mathrm{dtwd}$, calculated by means of the dynamic time warping with duration (DTWD) metric. The training's objective is to minimize the negative expected reward, which is equivalent to maximizing a positive reward:
\begin{flalign}
\loss(\theta)=-\mathbb{E}_{\mathbf{\hat{p}} \sim \pi_\theta} r(\mathbf{\hat{p}}),
\end{flalign}
where $\mathbf{\hat{p}} = (\hat{p}_1, ..., \hat{p}_T)$ and where $\hat{p}_i$ represents the $i$-th
fixation sampled from the model-generated distribution. The reward function combines the DTWD metric, assessing the similarity between the predicted and the ground truth scanpath, with the summed salient-value reward for each fixation point along the generated scanpath:  
\begin{flalign}
r(\mathbf{\hat{p}})= -r_{\mathrm{dtwd}}(\mathbf{\hat{p}}) + \sum_{i=1}^T r_\mathrm{sal}(\hat{p}_i).
\end{flalign}

\subsubsection{{Dynamic Time Warping with Duration (DTWD)}}
Dynamic Time Warping (DTW) is widely used for comparing two sequences that may differ in length \cite{berndt1994using, salvador2007toward}. It is useful for scanpaths because it finds an optimal alignment between the two scanpaths (ground truth and predicted ones) 
and computes the distance without missing any critical features. 
We implement DTWD as an extended DTW to consider both the spatial and temporal characteristics of scanpaths.
Specifically, we spatially align scanpaths using fixation positions, then compute DTWD value as 3D vectors $(x, y, t)$ of fixation position and duration. 
By incorporating DTWD computed over the full scanpath into the reward function, we sought to generate scanpaths closer to the ground truth trajectories and duration. 

%The optimal alignment is computed based on position, and then the DTW reward is calculated to account for duration as well.

\subsubsection{{Salient Value}}
EyeFormer applies rewards for salient values to encourage fixations in salient areas. 
To avoid repeatedly fixating on the same location in the image, 
we employ an inhibition-of-return mechanism~\cite{itti1998model} to model the relevant tendency of the human visual system.
We established inhibition areas on the saliency map for all the previously predicted fixation points. 
If the new predicted point falls within these areas, it does not evoke any additional reward.
\ADD{Note that predicted scanpaths can still return to an already-visited element, since 
DTWD encourages fixations to revisit the most salient areas 
and our chosen IOR radius is not so large as to preclude revisiting an element, as explained next.}
First, we denote the dimension of a display by $W \times H$. 
Then we set the diameter of the display's inhibition areas $m_{\textrm{display}}$ to be commensurate with human's visual angle, 
the angle an object subtends at the eye (see \autoref{fig:ior}a). 
Such diameter setting was suggested by \citet{KLEIN20011757} 
and has been further analyzed by \citet{Emami24_etra}.
Finally, we compute $m_{\textrm{orig}}$, the diameter for the corresponding inhibition areas for the input image with size $w_\image \times h_\image$ (see \autoref{fig:ior}b):
\begin{flalign}
m_{\textrm{orig}} = \dfrac{m_{\textrm{display}}}{\min\big(W / w_\image, H / h_\image\big)}.
\end{flalign}
Note that preparing the image for processing necessitates resizing it to $w_\textrm{inp} \times h_\textrm{inp}$, 
which corresponds to the input image size required by the policy model before splitting the input image into patches. 
The required input image is squared to ease computations~\cite{Dosovitskiy2020ViT}. 
Accordingly, we resize the inhibition areas from circles to ellipses, accounting for potential distortions (see \autoref{fig:ior}c). 
Any point $(x,y)$ in the image that satisfies the following condition gets inhibited (resulting in a salient-value reward of 0) 
and is omitted from the saliency map:
\begin{flalign}
\dfrac{(x-x_i)^2}{m_\textrm{w}^2} + \dfrac{(y-y_i)^2}{m_\textrm{h}^2} \leq 1, \ \text{where}\ &m_\textrm{w} = \dfrac{ w_\textrm{inp}}{w_\image}  m_{\textrm{orig}}, \ m_\textrm{h} = \dfrac{ h_\textrm{inp}}{h_\image}  m_{\textrm{orig}},
\end{flalign}
where $(x_i, y_i)$ are the coordinates for the $i$-th predicted fixation point.
Hence, salient-value reward $r_\mathrm{sal}$ at step $i$ is defined as the salient value of predicted fixation $\hat{p}_i$ on the saliency map with IOR applied.

\subsection{Policy Network}
A two-stage approach characterizes the policy network for scanpath prediction. The visual representation of any image $\image$ is learned through the image encoder ($\encoder$), after which a scanpath gets generated by means of the fixation decoder ($\decoder$). For population-level scanpath prediction, the visual embedding $\encoder(\image)$ is taken as input to the decoder. For individual-level prediction, feeding the decoder this input along with a viewer embedding, $e_u$, allows the model to generate personalized scanpaths for separate viewers.

\subsubsection{{Vision Encoder.}}
We use a Vision Transformer~\cite{ViT2021} as the vision encoder.
Specifically, the image is resized to a resolution of $w_\textrm{inp}\times h_\textrm{inp}$ and split into $n_{\image}$ non-overlapping patches for the vision encoder.
\ADD{Splitting functions mainly to speed up model inference, capture local information, and obtain global information from relationships between patches.}
Next, a linear projection, a convolution layer, is applied to convert these patches into single-dimension embeddings $e_{\image}^k\in \mathbb{R}^{d_{\image}}$ thus: 
\begin{flalign}
e_{\tilde{\image}}=[e_{\image}^{CLS}, e_{\image}^{1}, \dots, e_{\image}^{n_{\image}}] + e_{pos},
% + \{e_{u,j}\}_{n_{\image}+1} \odot  \mathbb{1}_u,
\end{flalign}
%%n
where $e_{\image}^{CLS}$ is a learnable vector for the image context, $[e_{\image}^{CLS}, e_{\image}^{1}, \dots, e_{\image}^{n_{\image}}]$ is a matrix concatenated from the vectors $e_{\image}^{CLS}, e_{\image}^{1}, \dots, e_{\image}^{n_{\image}}$, and $e_{pos}\in \mathbb{R}^{d_{\image} \times (n_{\image}+1)}$ is the positional matrix reflecting the position context of the image patches. 
Finally, applying a vision encoder $\encoder(\cdot)$ based on a 12-layer version of the Vision Transformer (ViT) model~\cite{Dosovitskiy2020ViT}. ViT applies convolution per patch and uses a transformer to combine patch embeddings. It thus expresses the relationship for each patch and lets us derive the final image embedding, denoted as $\encoder(e_{\tilde{\image}})$.
We add an additional comparison between employing Vision Transformer and ResNet as the vision encoder in the Supplementary Materials.

\subsubsection{{Fixation Decoder.}}

To generate fixation points, $\hat{p}_{i}$, we use a multi-layer Transformer decoder, $\decoder$. It takes the image embedding $\encoder(e_{\tilde{\image}})$ alongside the previously generated points denoted by $\hat{p}_{1:i-1}$ as input to generate $\decoder(\encoder(e_{\tilde{\image}}), \hat{p}_{1:i-1})$. It allows previous fixation points to influence subsequent points in the scanpath.
We set the first fixation to be at the center of the screen since most eye-tracking datasets were collected by asking participants to look at the center of the images before seeing the images~\cite{jiang2023ueyes, xu2014predicting}. 
For the given state (the previously predicted fixation points and input image), the action (the next prediction of the fixation point) is 
\begin{flalign}
\pi_{\theta}(\hat{p}_{1:T}| \image) =\pi_{\theta}(\hat{p}_1|\image)\prod_{i=2}^{T} \pi_{\theta}(\hat{p}_{i} | \hat{p}_{1:i-1}, \image).
\end{flalign}
\Yue{
The policy $\pi_{\theta}$ is represented as a Gaussian distribution $\normal(\mu_i, \sigma_i)$. 
Alternatively, it can be represented as a mixed Gaussian distribution $\sum\limits_{k=1}^K \lambda_{ik} \normal(\mu_{ik}, \Sigma_{ik})$ with a total of $K$ Gaussian components, where $\lambda_{ik}$ denotes the weight of the $k$-{th} Gaussian component, and $\Sigma_{ik}$ denotes the specific covariance matrix specific for that component at step $i$. 
These variables for determining the distribution are sequentially generated by the decoder. 
We provide more implementation details and the comparison between the use of the Gaussian distribution and the mixed Gaussian distribution in the Supplementary Materials.}

\subsection{Predicting Personalized Scanpaths}

To distinguish between individual viewers, we leverage a two-layer Transformer architecture as the viewer encoder $\encoder_u$, facilitating the prediction of individual-level scanpaths.
In the training process, we train the model using the training users in the dataset. 
The viewer encoder is trained to allow each viewer's distinct viewing behaviors to be encoded in a separate embedding space.
In the test process, given a new viewer, the model updates the viewer encoder with a few scanpaths from that viewer by backpropagating from the scanpath samples.
Once the model has updated the viewer encoder, it can predict scanpaths specific to this unique viewer, thereby customizing its predictions for this individual's viewing behaviors.

Specifically, the image representation, $\encoder(e_{\tilde{\image}})$, serves as the input query, while the viewer embedding $e_{u}$ acts as the key and value in the cross-attention mechanism within the viewer encoder. 
The viewer embedding is a learnable matrix.
The output of the viewer encoder, $\encoder_u(e_{\tilde{\image}}, e_{u})$, is directed to the fixation decoder for the generation of fixations. 
The viewer encoder is not applied to the population-level predictions.

\subsection{Policy Gradient}

\ADD{To compute the gradient of the objective function $\nabla_\theta \loss(\theta)$, 
our method employs the REINFORCE algorithm~\cite{williams1992simple, rennie2017self},
which offers a Monte Carlo variant of a policy-optimization technique commonly used in RL~\cite{sutton2018reinforcement}.   
Under this algorithm, the agent accumulates samples from episodes by executing its current policy and utilizes those samples to update the policy's parameters iteratively.
The REINFORCE algorithm aims to maximize the cumulative expected reward across sequential actions by approximating the gradient of the expected reward for the policy's parameters. By adjusting these parameters iteratively in accordance with the gradient estimate, the algorithm attempts to enhance the policy's performance over time.}
This algorithm is rooted in the insight that the expected gradient of a non-differentiable reward function can be calculated as follows:
\begin{flalign}
\nabla_\theta \loss(\theta) = -\mathbb{E}_{\mathbf{\hat{p}} \sim \pi_\theta} \left[r(\mathbf{\hat{p}})\nabla_\theta \log \pi_\theta(\mathbf{\hat{p}} | \image)\right].
\end{flalign}

To approximate the expected gradient, we use a single Monte-Carlo sample $\mathbf{\hat{p}} = (\hat{p}_1, ..., \hat{p}_T)$ from the policy $\pi_\theta$ for each training example in the minibatch:
\begin{flalign}
\nabla_\theta \loss(\theta) \approx -r(\mathbf{\hat{p}})\nabla_\theta \log \pi_\theta(\mathbf{\hat{p}} | \image).
\end{flalign}

\paragraph{\textbf{REINFORCE with a Baseline.}} 

Our technique uses a baseline $b$ to assess the environment's expected reward without any actions, thus generalizing the policy gradient obtained from REINFORCE. Applying this algorithm with a baseline allows us to estimate the advantage yielded by an action -- i.e., the difference between the actual reward obtained and that expected from the baseline environment. By subtracting the baseline value, we reduce the variance of the gradient estimation, thus arriving at a more stable optimization process. The gradient of the loss with respect to the $\theta$ policy parameters is then obtained as
\begin{flalign}
\nabla_\theta \loss(\theta) = -\mathbb{E}_{\mathbf{\hat{p}} \sim \pi_\theta} \left[(r(\mathbf{\hat{p}}) - b) \nabla_\theta \log \pi_\theta(\mathbf{\hat{p}} | \image)\right].
\end{flalign}

For each step in the training, our technique approximates the expected gradient with a single sample $\mathbf{\hat{p}} \sim \pi_\theta$:
\begin{flalign}
\nabla_\theta \loss(\theta) \approx -(r(\mathbf{\hat{p}}) - b)\nabla_\theta \log \pi_\theta(\mathbf{\hat{p}} | \image).
\end{flalign}

In the discrete space,  Rennie et al.'s conceptualization~\cite{rennie2017self} serves as a foundational framework, 
where $b$ is estimated by the reward obtained from the greedy search of policy.
However, the approach diverges when operating in a continuous space. At each step, the operation of our policy necessitates the computation of $b$, determined as the reward associated with the mean of multiple samples drawn from the policy --- essentially, the mean of the distribution generated by the policy.
Consequently, the expected gradient is calculated as
\begin{flalign}
\nabla_\theta \loss(\theta) \approx -(r(\mathbf{\hat{p}}) - r(sg[\boldsymbol{\mu}]))\nabla_\theta \log \pi_\theta(\mathbf{\hat{p}} | \image),
\end{flalign}
where $\boldsymbol{\mu}=(\mu_1,\dots, \mu_T)$ and $sg[\cdot]$ stands for a stop-gradient operator having zero partial derivatives.

\def\arraystretch{1}% 
\begin{table*}[t!]
\scalebox{0.82}{\setlength\tabcolsep{5pt}
\begin{tabular}{lcccccccccc}
\hline
\multirow{ 2}{*}{\textbf{Model}}  & \multirow{ 2}{*}{\textbf{DTW} $\downarrow$}  & \multirow{ 2}{*}{\textbf{TDE} $\downarrow$} & \multirow{ 2}{*}{\textbf{Eyenalysis} $\downarrow$} & \multirow{ 2}{*}{\textbf{DTWD} $\downarrow$} & \multicolumn{6}{|c}{\textbf{MultiMatch} $\uparrow$}  \\
 &  &  &   & &  \multicolumn{1}{|c}{\textbf{Shape}}   & \textbf{Direction}  &  \textbf{Length}   & \textbf{Position}   & \textbf{Duration}  & \textbf{Mean}  \\
\hline
 Personalized to Other Viewers &  4.152  $\pm$ 1.161  & 0.123  $\pm$ 0.030 &   \bf  0.036  $\pm$ 0.017 &  5.070  $\pm$ 1.088 &   \bf 0.943 & 0.733 & 0.935 &  0.821 &   \bf 0.731 & 0.833  \\
%\hline
%\bf Population-Level &   4.082 $\pm$  1.124    &  \bf 0.121  $\pm$  0.028   &   \bf 0.035  $\pm$  0.017   &    5.054  $\pm$  1.088 &   0.942    &  \bf 0.748    &  \bf  0.940   &   0.825   &    0.750   &  0.841    \\
\hline
Personalized to Target Viewer &   \bf  4.058 $\pm$  1.135   &   \bf 0.121  $\pm$  0.029   &    \bf 0.036  $\pm$  0.017   & \bf   4.996  $\pm$  1.078 &  \bf  0.943    &   \bf  0.737   &   \bf  0.936   &    \bf  0.824   &   \bf   \bf 0.731   & \bf     \bf 0.834    \\
\bottomrule
\end{tabular}
}
\caption{
    We compare the model personalized to the target test viewer against the model personalized to other test viewers to quantify the effectiveness in capturing the characteristics of individual viewers on the UEyes dataset.
}
\Description{A table showing the comparison between the model personalized to the target test viewer and the model personalized to other test viewers.}
\label{tbl:individual_table}
\end{table*}

\setlength{\tabcolsep}{0.9pt}
\def\arraystretch{0.2}% 
\begin{figure*}[t!]
\def\w{0.140\linewidth}
 \centering
 \begin{tabular}{cccccccccccc}
 & Viewer 1 & Viewer 2 & Viewer 1  & Viewer 2 & Viewer 1  & Viewer 2  \\
 \begin{turn}{90} 
\ \ \ \ \ \ \ \  \ \ \ \ \ \ GT
\end{turn} & 
\includegraphics[height=\w]{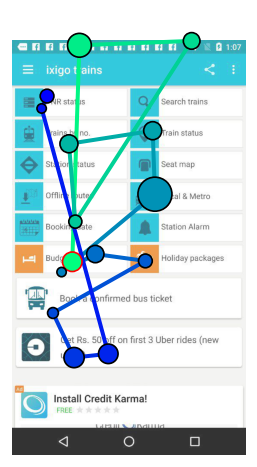} &
\includegraphics[height=\w]{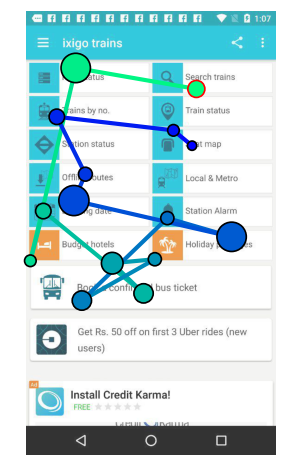} &

\includegraphics[height=\w]{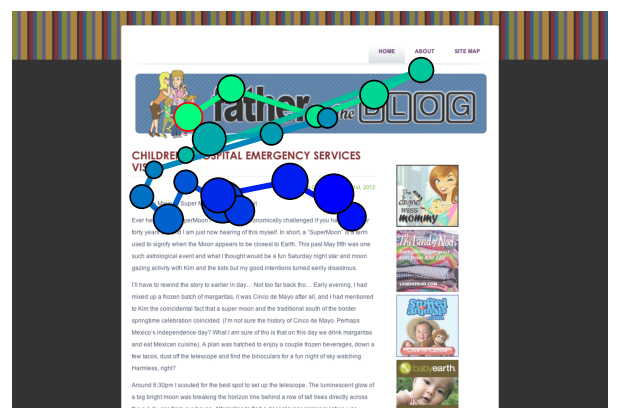} &
 \includegraphics[height=\w]{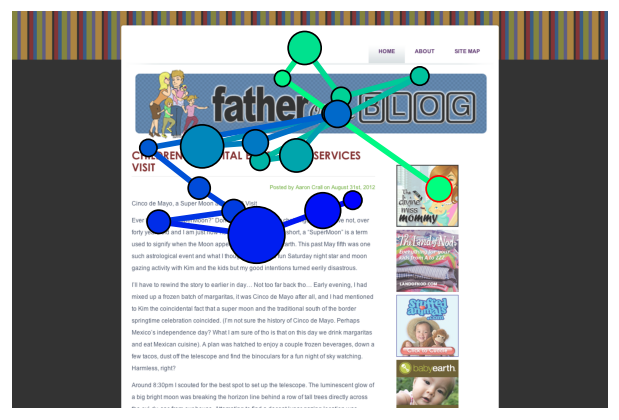} &
\includegraphics[height=\w]{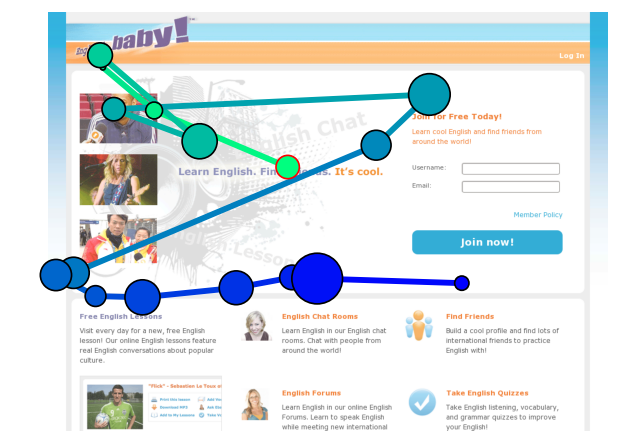} &
\includegraphics[height=\w]{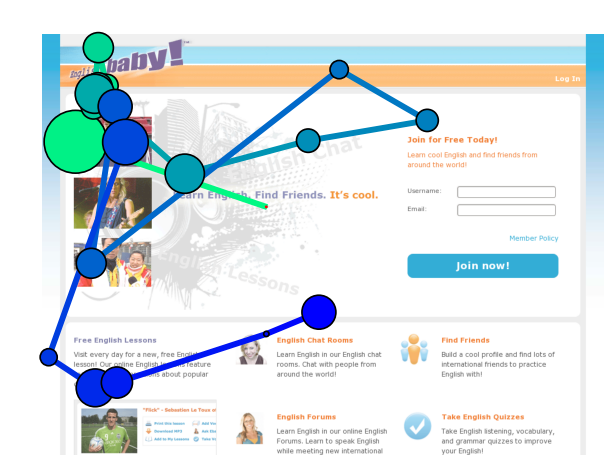}
\\
 \begin{turn}{90} 
\ \ \ \  \ \ \ \ Prediction
\end{turn} & 
\includegraphics[height=\w]{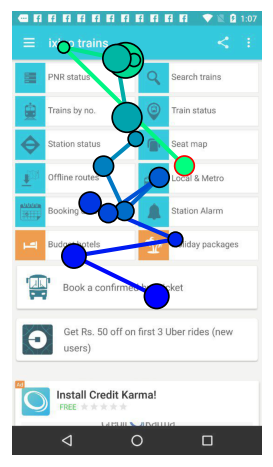} & 
\includegraphics[height=\w]{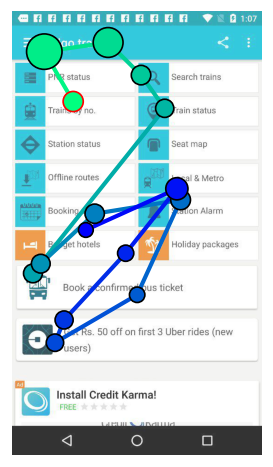} &

\includegraphics[height=\w]{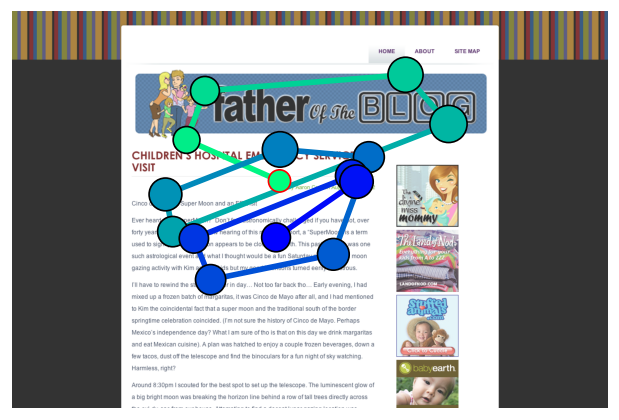}   &
\includegraphics[height=\w]{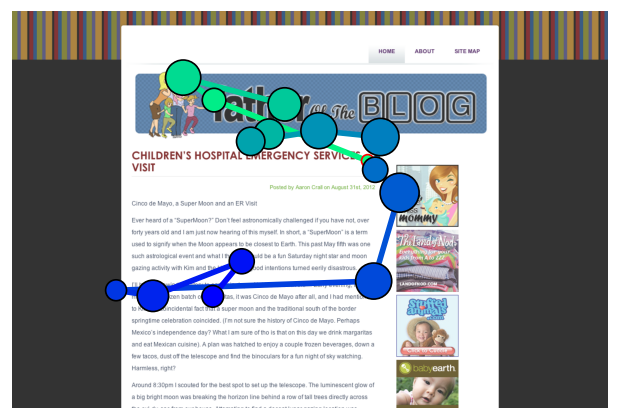} & 

\includegraphics[height=\w]{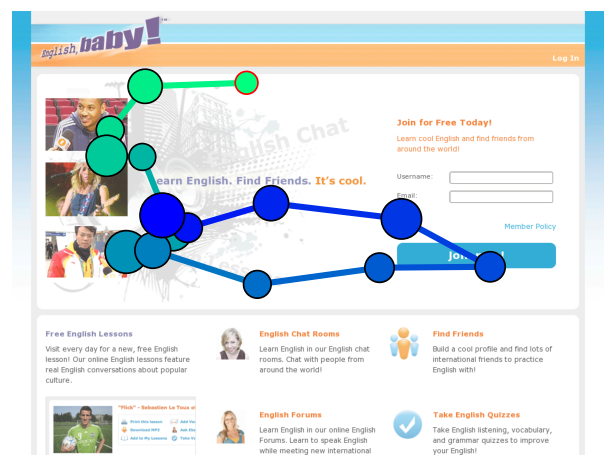}  &
\includegraphics[height=\w]{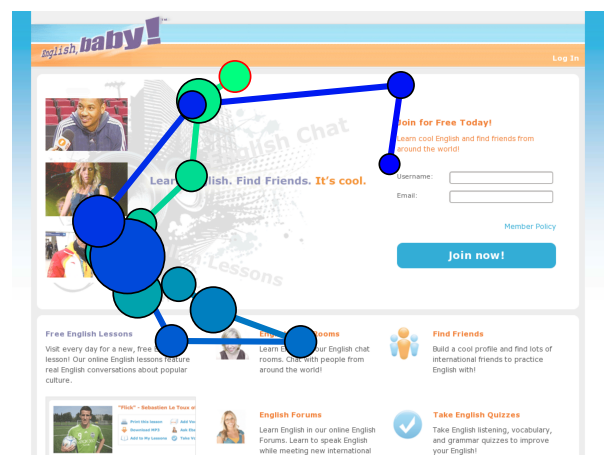}  &
\includegraphics[height=\w]{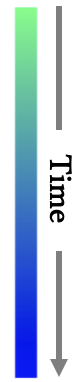}
\\
\end{tabular}

\def\w{0.098 \linewidth}
 \centering
\begin{tabular}{*8c}
 & Viewer 1 & Viewer 2 & Viewer 1 & Viewer 2 & Viewer 1 & Viewer 2 \\
 \begin{turn}{90} 
\ \ \ \  \ \ \  \ \ GT
\end{turn} & 
\includegraphics[height=\w]{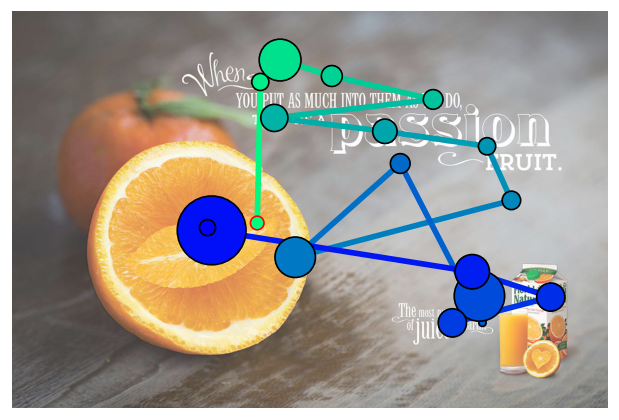} &
\includegraphics[height=\w]{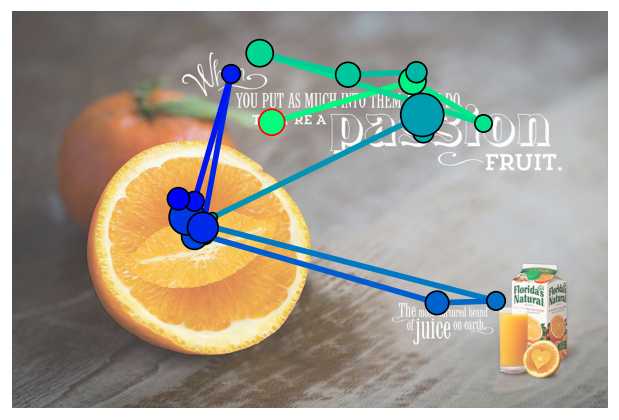} &
\includegraphics[height=\w]{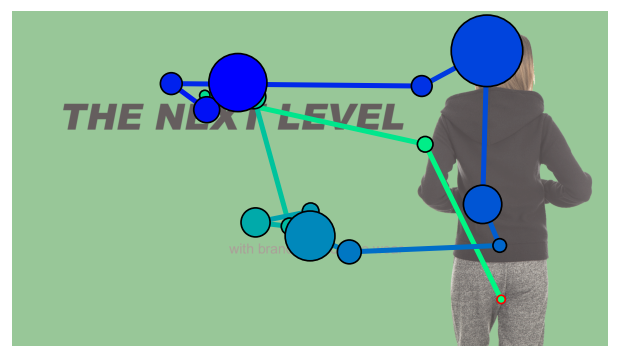} &
\includegraphics[height=\w]{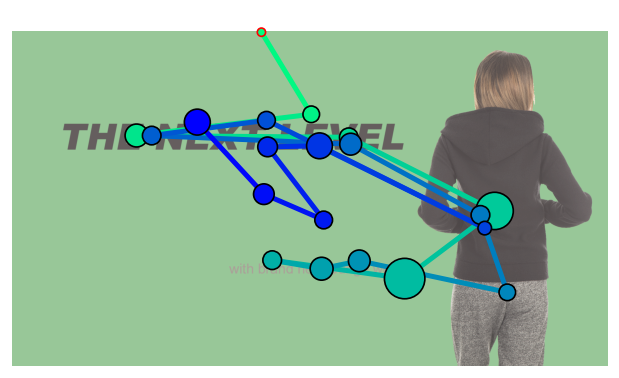} &
\includegraphics[height=\w]{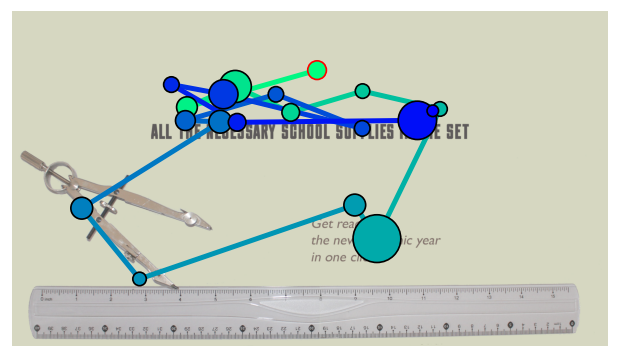} &
\includegraphics[height=\w]{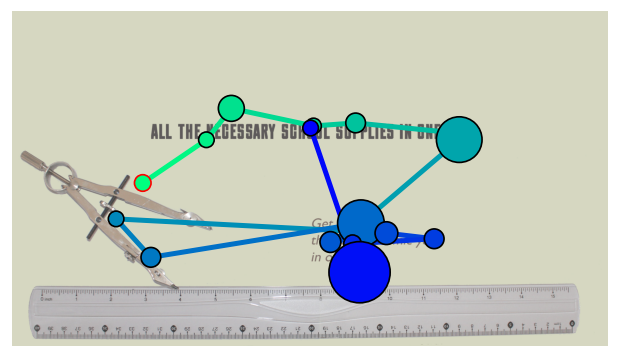} 

\\
 \begin{turn}{90} 
 \ \ \ Prediction
\end{turn} & 
\includegraphics[height=\w]{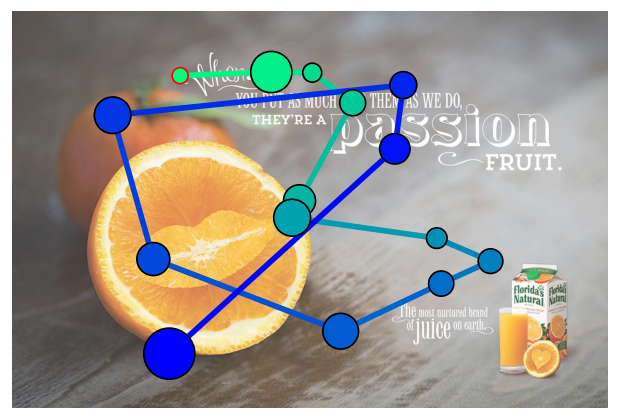} &
\includegraphics[height=\w]{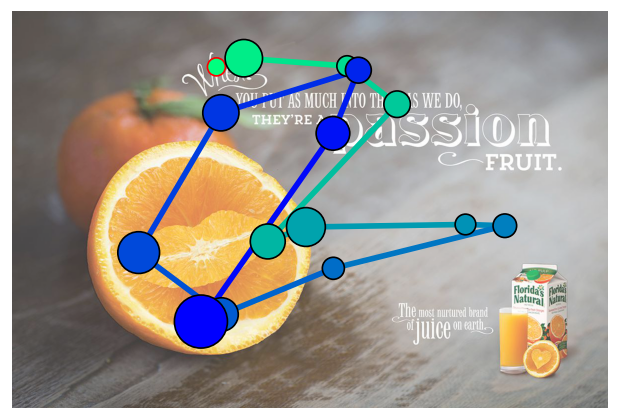} &
\includegraphics[height=\w]{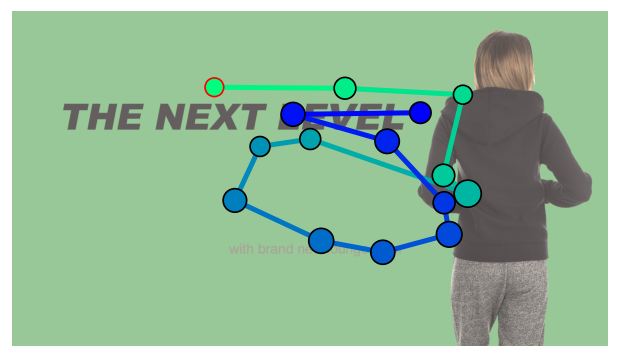} &
\includegraphics[height=\w]{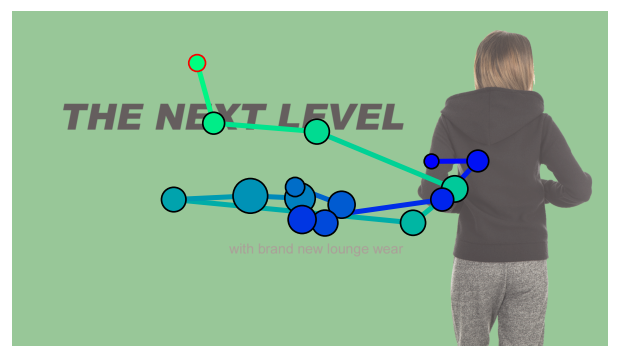} &
\includegraphics[height=\w]{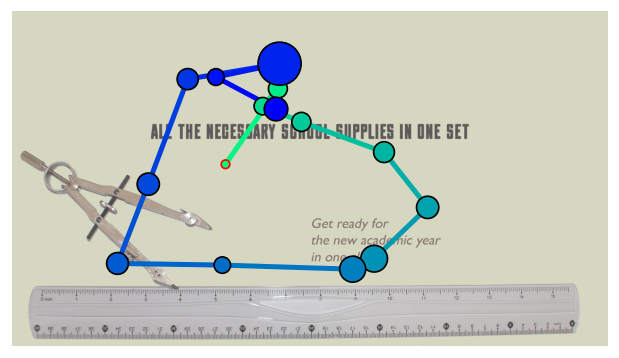} &
\includegraphics[height=\w]{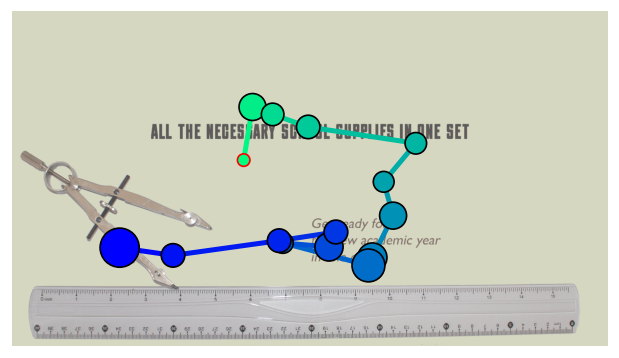} &
\includegraphics[height=\w]{source/figures/colormap.png}
\end{tabular}
\caption{
%By leveraging a few scanpath samples from each viewer, our model demonstrates the ability to generate personalized scanpaths. Note that we use Viewer 1 and Viewer 2 to show the prediction of two viewers on the same images, but they are not always the same viewers in all the examples.
Scanpaths personalized for two viewers, illustrating our model's ability to generate these by means of only a few scanpath samples from each viewer (note that ``Viewer 1'' and ``Viewer 2'' are generic terms; the viewers are not the same across all examples). More examples are presented in Supplementary Materials.
}
\Description{Results of scanpaths personalized for two viewers, illustrating our model's ability to generate these by means of only a few scanpath samples from each viewer.}
\label{fig:shortened_fewshot_results}
\end{figure*}

\setlength{\tabcolsep}{1.5pt}
\def\arraystretch{0.4}% 
\begin{figure*}[t!]
\def\w{0.153\linewidth}
 \centering
\begin{tabular}{c *6c}
\bf \begin{turn}{90} 
\bf \ \ \ \ \ \   Ground Truth
\end{turn} &
\includegraphics[height=\w]{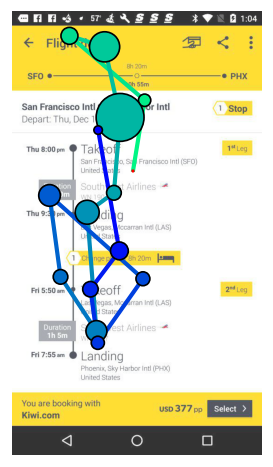} &
\includegraphics[height=\w]{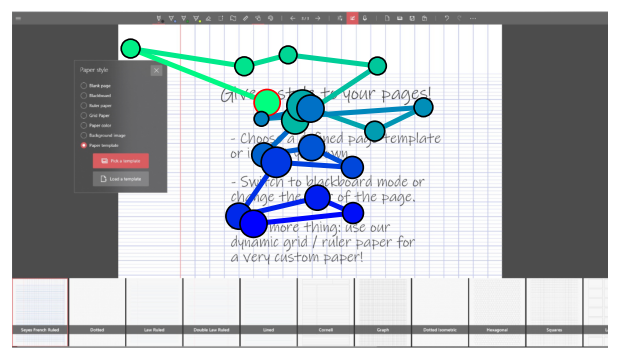} &
\includegraphics[height=\w]{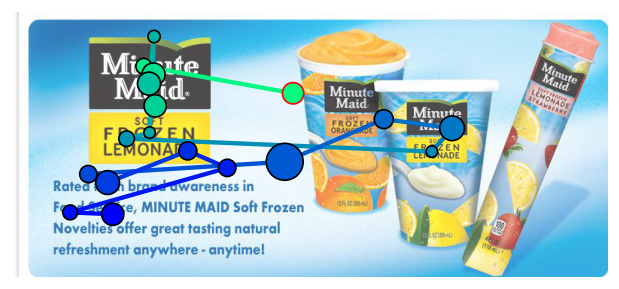} &
\includegraphics[height=\w]{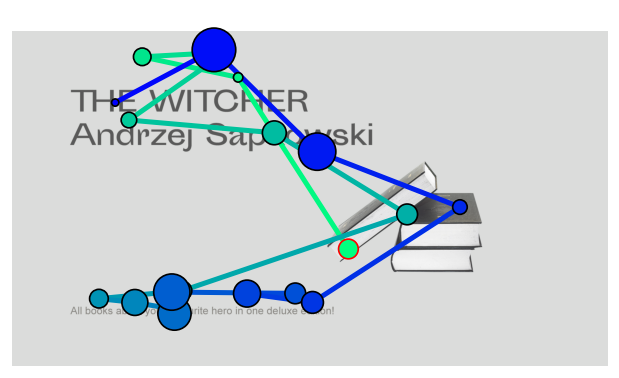} 
\\
\bf \begin{turn}{90} 
\bf \ \ \ \ \ \ \ \ \ \  Prediction
\end{turn} &
\includegraphics[height=\w]{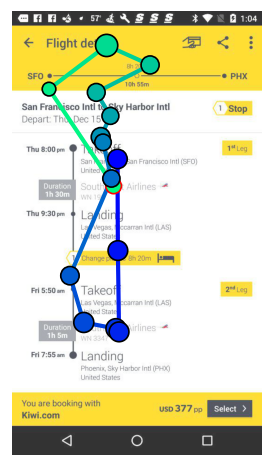} &
\includegraphics[height=\w]{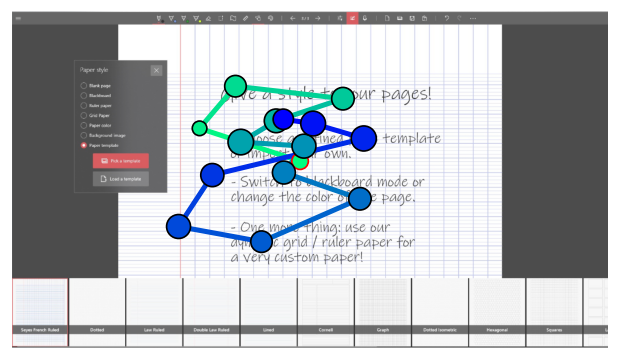}  &
\includegraphics[height=\w]{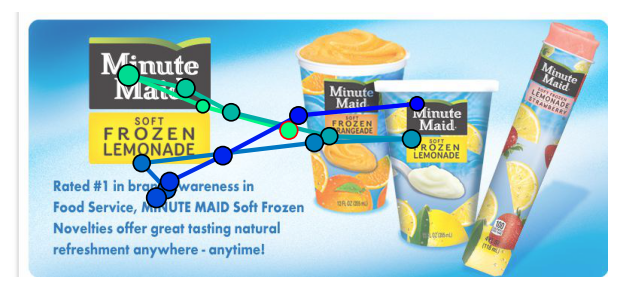} &
\includegraphics[height=\w, trim=20 0 30 10, clip]{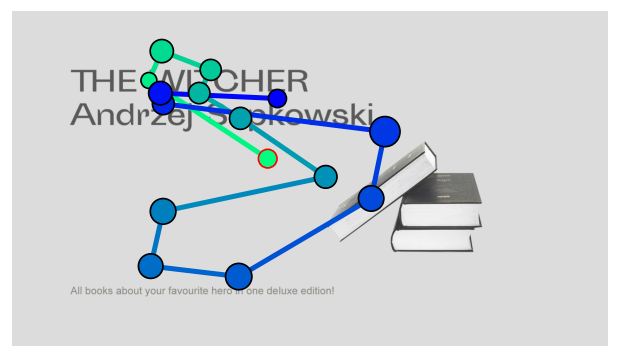}  &
\includegraphics[height=\w]{source/figures/colormap.png}
\\
\end{tabular}
\def\w{0.176\linewidth}
 \centering
\begin{tabular}{c *9c}
\bf \begin{turn}{90} 
\bf \ \ \ \ \ \ \ \ \    Ground Truth
\end{turn} &
\includegraphics[height=\w]{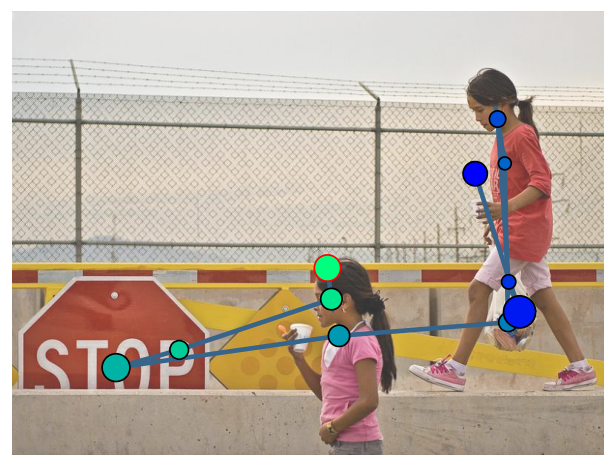} &
\includegraphics[height=\w]{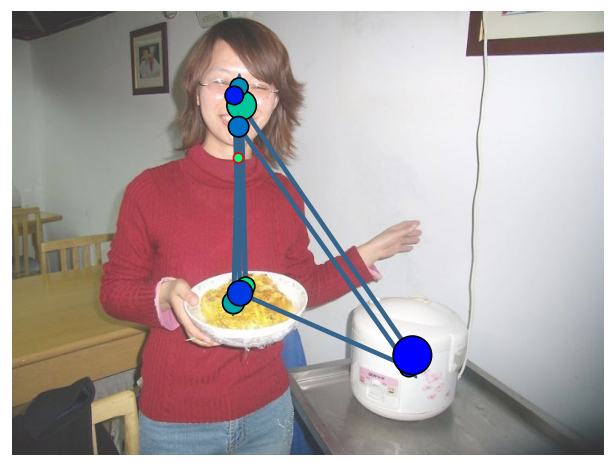} &
\includegraphics[height=\w]{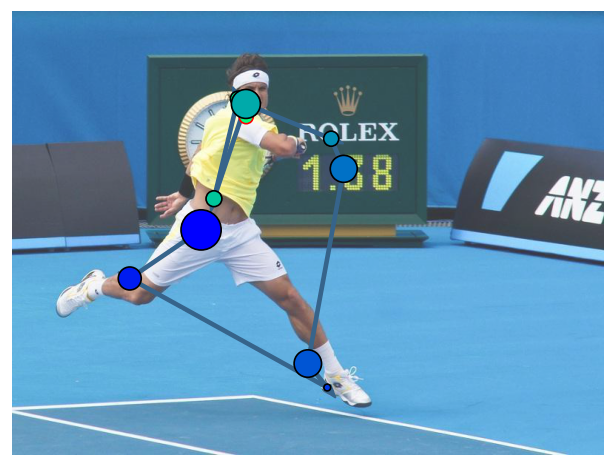} &
\includegraphics[height=\w]{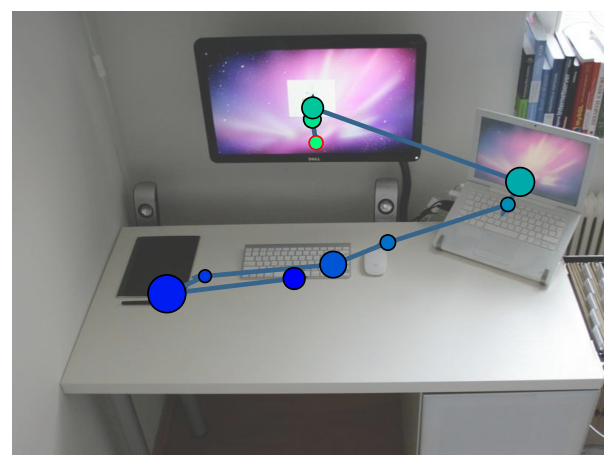} &
\\
\bf \begin{turn}{90} 
\bf \ \ \ \  \ \ \ \ \ \ \ \ \  Prediction 
\end{turn} &
\includegraphics[height=\w]{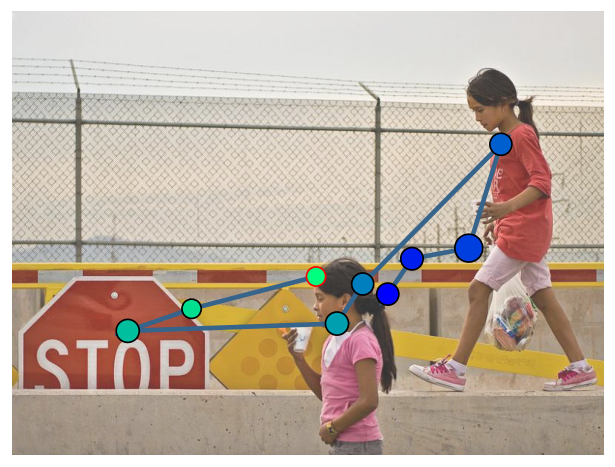} &
\includegraphics[height=\w]{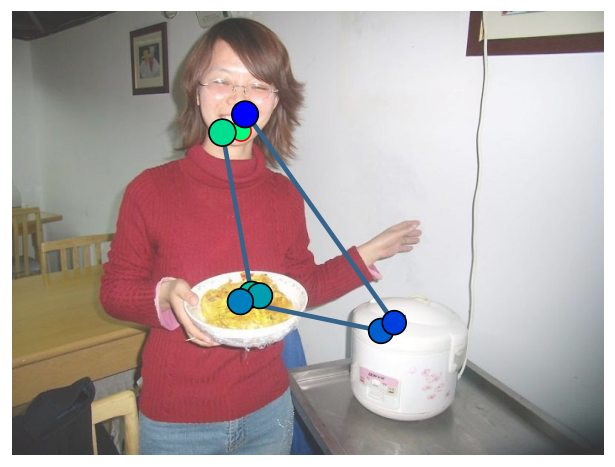} &
\includegraphics[height=\w]{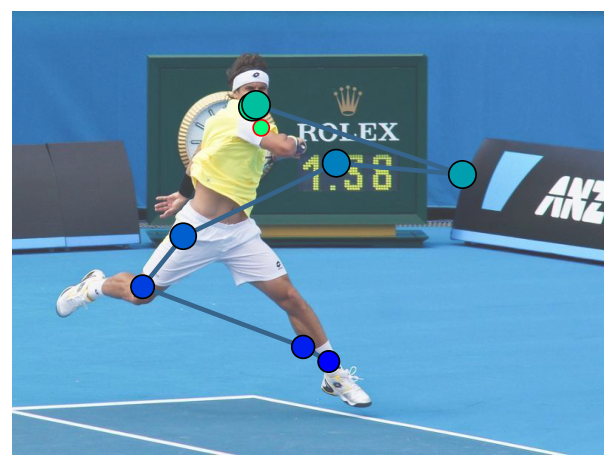} &
\includegraphics[height=\w]{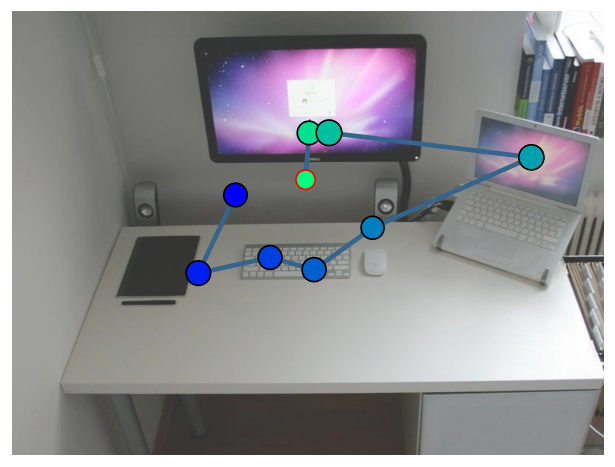} &
\includegraphics[height=\w]{source/figures/colormap.png}
\\
\end{tabular}
\vspace{-4.5mm}
\caption{
   Our population-level scanpath prediction shows that they are close to the ground truth (GT) regarding fixation positions, ordering, and duration. More examples are presented in Supplementary Materials.
}
\Description{Results of our population-level scanpath prediction.}
\label{fig:shortened_scanpath_results}
\end{figure*}

\setlength{\tabcolsep}{2pt}
\def\arraystretch{1}% 
\begin{figure*}[t!]
  \def\w{0.96\linewidth}
  \centering
 \includegraphics[width=\w]{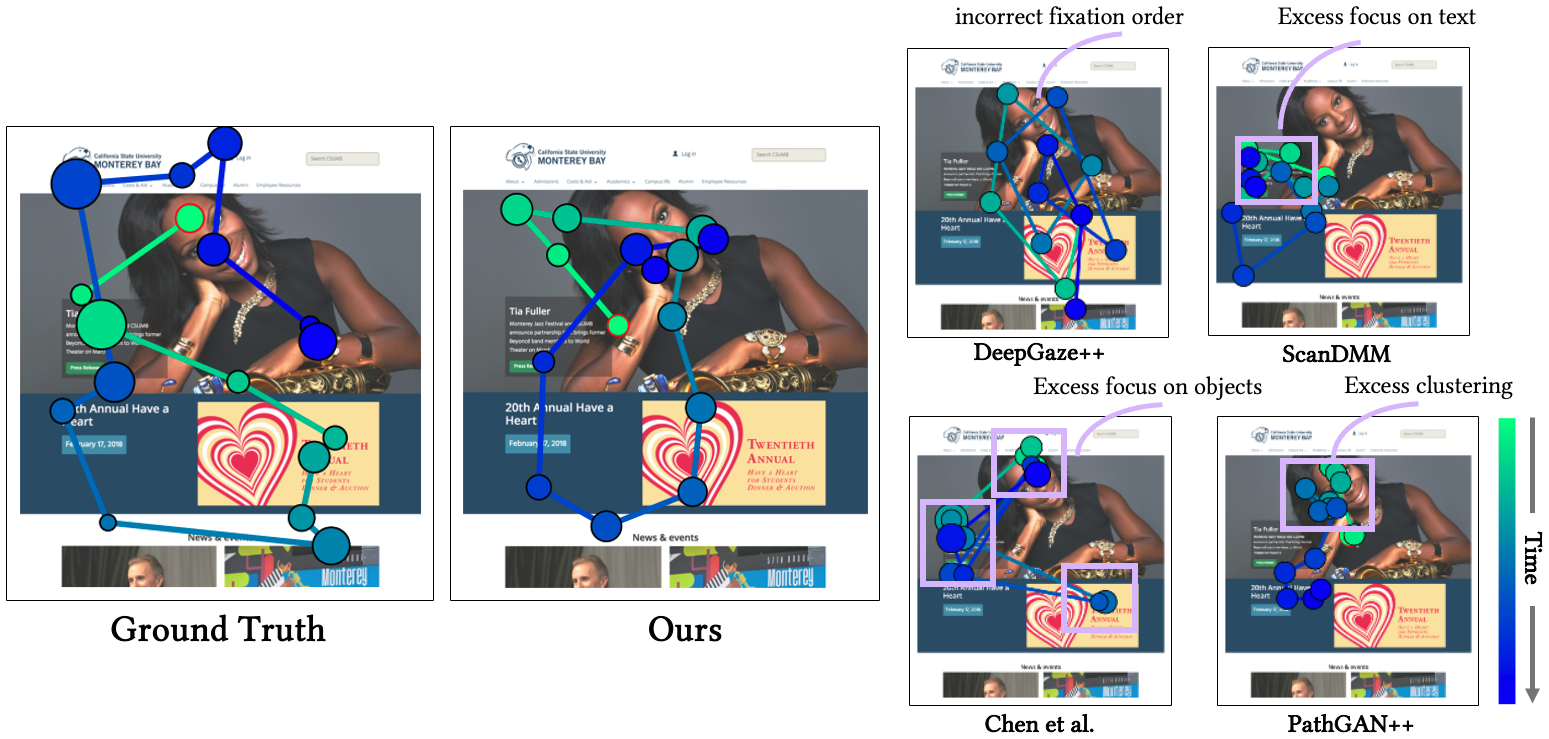}
\caption{
    \ADD{Annotated comparison between different models.
    This illustration presents the best baseline methods, with annotated limitations.}
}
\Description{Annotated comparison between different models. This illustration presents the best baseline methods, with annotated limitations.}
\label{fig:shortened_scanpath_qualitative}
\end{figure*}

\setlength{\tabcolsep}{2pt}
\def\arraystretch{1}% 
\begin{figure*}[t!]
\def\w{0.185\linewidth}
\def\ww{0.155\linewidth}
 \centering
\begin{tabular}{c *9c}
\bf \begin{turn}{90} 
\bf \ \ \ \ \ \ \ \REVISION{Ground Truth}
\end{turn}  &
\includegraphics[height=\w, trim=10 10 10 10, clip]{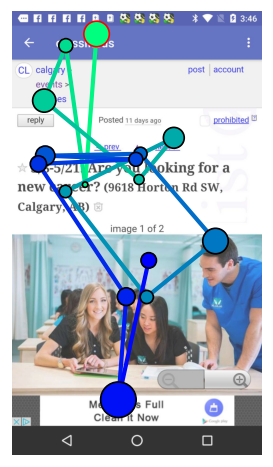} &
\includegraphics[height=\w, trim=16 20 16 10, clip]{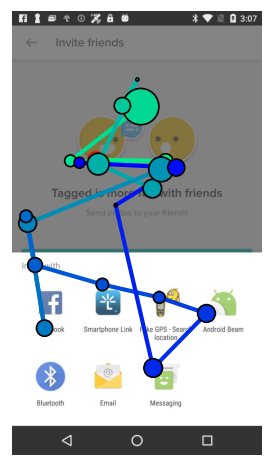} &
\includegraphics[height=\w, trim=10 10 10 5, clip]{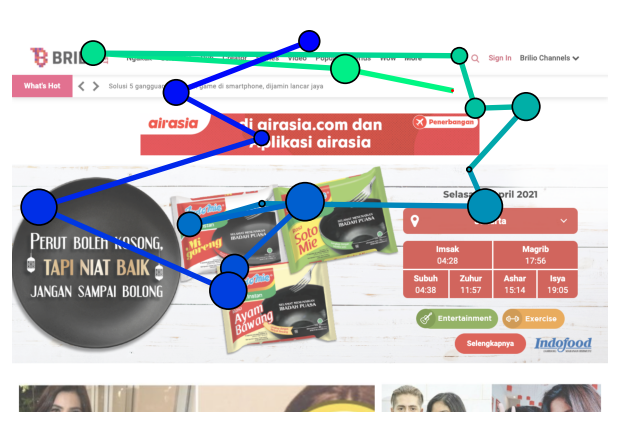} &
\includegraphics[height=\w, trim=18 20 18 18, clip]{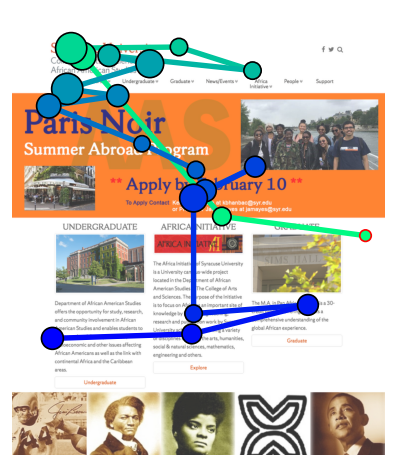} & 
\includegraphics[height=\w, trim=10 10 10 10, clip]{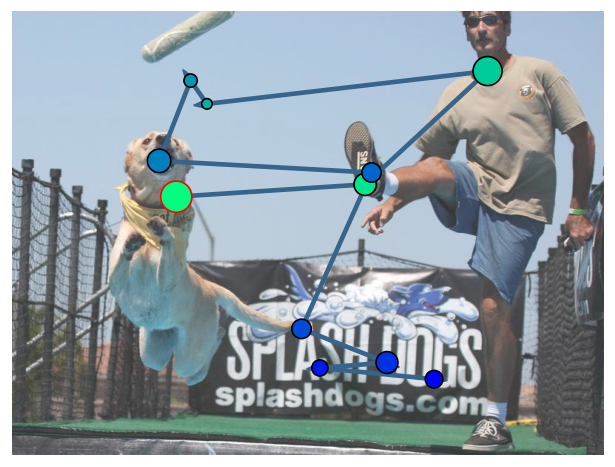} 
\\
\bf \begin{turn}{90} 
\ \ \ \ \ \ \ \ \ \   \REVISION{\begin{tabular}{@{}c@{}} Chen et al. \end{tabular}}
\end{turn}  &
\includegraphics[height=\w, trim=10 10 10 10, clip]{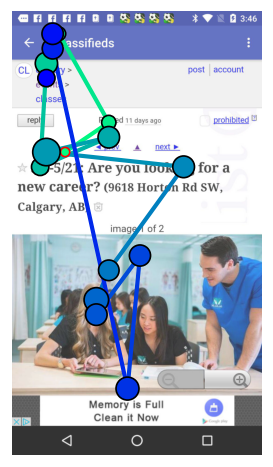}&
\includegraphics[height=\w, trim=16 20 16 10, clip]{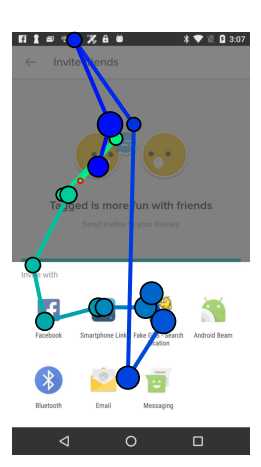}&
\includegraphics[height=\w, trim=10 10 10 5, clip]{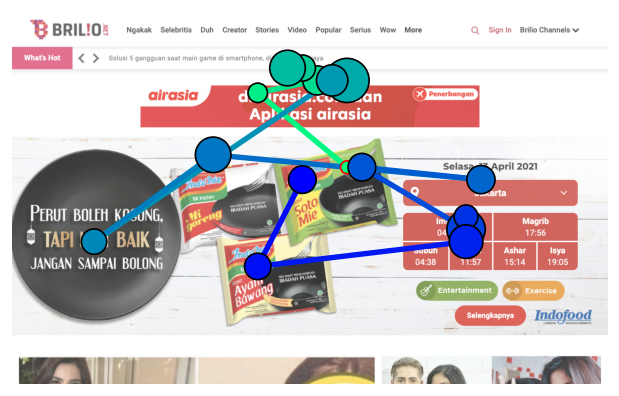}&
\includegraphics[height=\w, trim=18 20 18 15, clip]{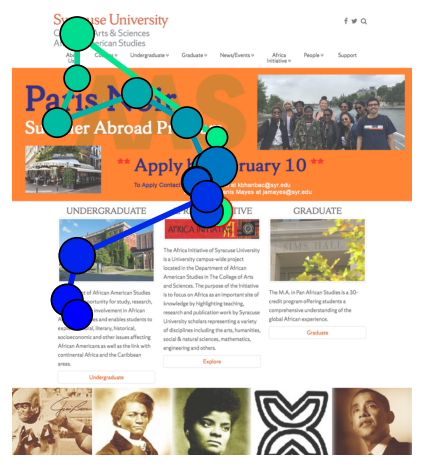} & 
\includegraphics[height=\w, trim=10 10 10 10, clip]{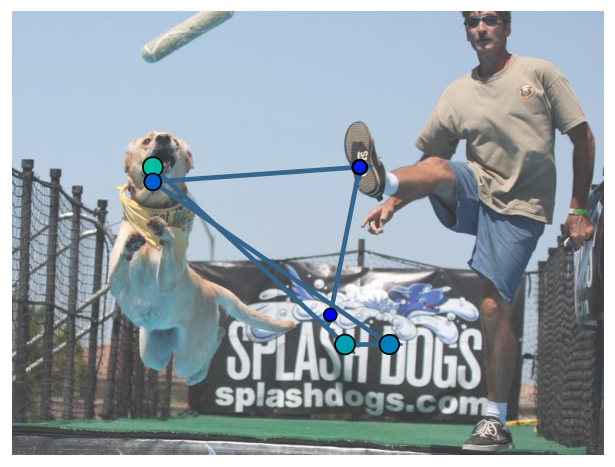}
\\
\bf \begin{turn}{90} 
\bf \ \ \ \ \ \ \ \ \ \ \ \ \ \ \ \REVISION{\begin{tabular}{@{}c@{}}Ours\end{tabular}}
\end{turn} &
\includegraphics[height=\w, trim=10 10 10 10, clip]{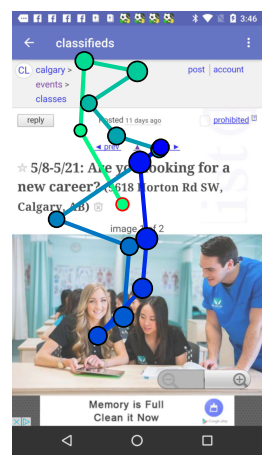} &
\includegraphics[height=\w, trim=16 20 16 10, clip]{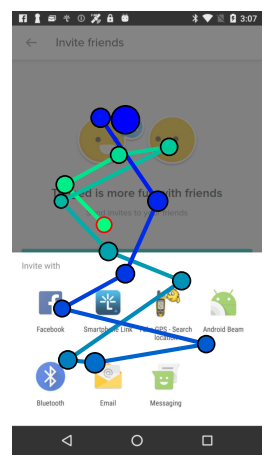} &
\includegraphics[height=\w, trim=10 10 10 5, clip]{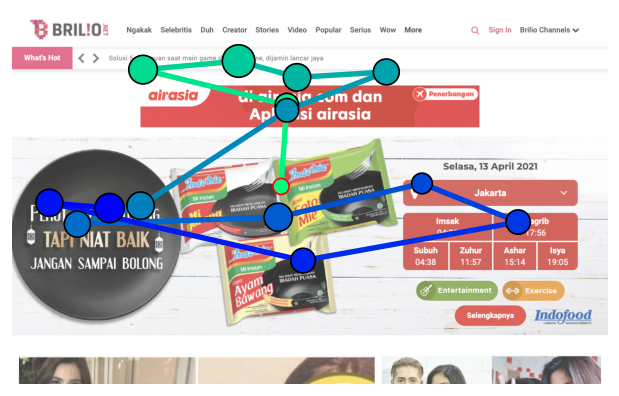} &
\includegraphics[height=\w, trim=18 20 18 15, clip]{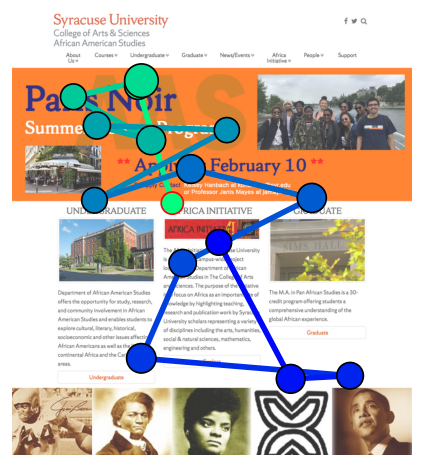} & 
\includegraphics[height=\w, trim=10 10 10 10, clip]{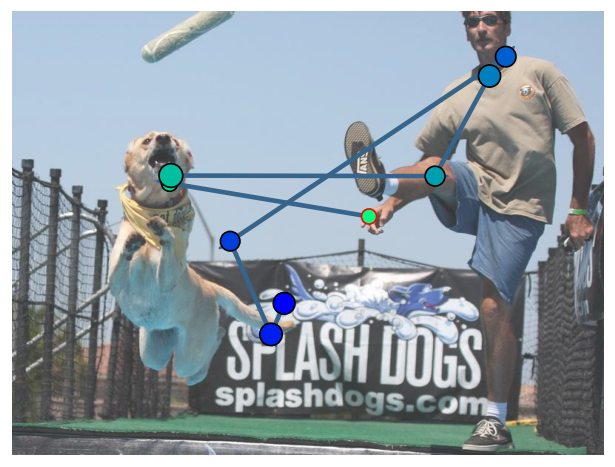}  &
\includegraphics[height=\w]{source/figures/colormap.png}
\\
\end{tabular}
\caption{
    \ADD{Qualitative comparison: This illustration presents the best baseline method~\citet{chen2021predicting}. We show the comparison to other scanpath models in the Supplementary Materials.}
}
\Description{Qualitative comparison: This illustration presents the best baseline method.}
\label{fig:scanpath_qualitative_chen}
\end{figure*}
\def\arraystretch{1}% 
\begin{table*}[t!]
\scalebox{0.85}{\setlength\tabcolsep{5pt}
\begin{tabular}{lcccccccccc}
\hline
\multirow{ 2}{*}{\textbf{Model}}  & \multirow{ 2}{*}{\textbf{DTW} $\downarrow$}  & \multirow{ 2}{*}{\textbf{TDE} $\downarrow$} & \multirow{ 2}{*}{\textbf{Eyenalysis} $\downarrow$} & \multirow{ 2}{*}{\textbf{DTWD} $\downarrow$} & \multicolumn{6}{|c}{\textbf{MultiMatch} $\uparrow$}  \\
 &  &  &   & &  \multicolumn{1}{|c}{\textbf{Shape}}   & \textbf{Direction}  &  \textbf{Length}   & \textbf{Position}   & \textbf{Duration}  & \textbf{Mean}  \\
\hline
\multicolumn{11}{c}{\textbf{GUIs (UEyes dataset)}} \\
\hline
Itti--Koch &   6.249 $\pm$ 0.986 &   0.150 $\pm$ 0.025 &     0.047 $\pm$ 0.028 &  --- &   0.861 &  0.721 & 0.819 &  0.746 &  ---&  --- \\ 
DeepGaze III Pretrained   &  7.906 $\pm$ 2.466  & 0.274 $\pm$ 0.061  &  0.124 $\pm$ 0.076 &  --- &  0.937 &  0.567 &  0.886&  0.746 &  ---&  --- \\ 
DeepGaze++ &  5.454  $\pm$ 1.078 & 0.149  $\pm$ 0.032 & 0.047  $\pm$ 0.026 &  --- &  0.907 & 0.708 &  0.906 &  0.773 &  ---&  --- \\ 
{PathGAN Pretrained}    &     4.719 $\pm$ 1.387 &    0.192 $\pm$ 0.049 &    0.072 $\pm$ 0.037&  ---  & 0.940 &  0.579 & 0.892 & 0.800   & --- & ---   \\ 
{PathGAN}         &  4.754 $\pm$ 1.185 &   0.147 $\pm$ 0.048 &   0.048 $\pm$ 0.025 &  ---  & \bf 0.943 & 0.716 & 0.935 & 0.797 & --- & ---  \\ 
{PathGAN++}   &     4.559 $\pm$ 1.182 &     0.146 $\pm$ 0.037 &   0.044 $\pm$ 0.022 &   --- & \bf 0.943 & 0.706 & 0.933 & 0.807 &  --- & --- \\ 
{PathGAN w/ D}     &  5.192 $\pm$ 1.422 &   0.204 $\pm$ 0.045 &   0.092 $\pm$ 0.038 & 6.431 $\pm$ 1.644 & 0.939 & 0.556 &  0.891 & 0.779 & 0.667 & 0.766 \\ 
{PathGAN++ w/ D}        &  5.443 $\pm$ 1.466 &   0.202 $\pm$ 0.044 &   0.096 $\pm$ 0.043 & 6.667 $\pm$ 1.659  & 0.939 & 0.560 & 0.896  & 0.765  & 0.657 & 0.763 \\ 
{SaltiNet}   &   7.042 $\pm$ 1.622 &   0.187 $\pm$ 0.057 &   0.063 $\pm$ 0.054 & 8.241 $\pm$ 1.487  &  0.907   &  0.715   &  0.897  &  0.691   &   0.579   &    0.758   \\
{UMSS}   &   5.051 $\pm$ 1.592 &   0.155 $\pm$ 0.048 &   0.050 $\pm$ 0.026 & 6.495 $\pm$ 1.468  &  0.934   &  0.713   &  0.921  &  0.779   &   0.579   &    0.785   \\
{ScanGAN}   &   4.815 $\pm$ 1.238   &   0.136 $\pm$ 0.034    &    0.040 $\pm$ 0.022 
  &   --      &   0.931    &    0.734   &   0.929    &    0.796    &   --   &  --          \\
{ScanDMM}   &   5.085 $\pm$ 1.317   &    0.138  $\pm$  0.037   &    0.043   $\pm$  0.027    &   --  &   0.931   &   0.729   &   0.928   &   0.784  &   --   &   --          \\
{Chen et al.}  &   4.335  $\pm$  1.299   &   \bf 0.118   $\pm$  0.034   &   0.037  $\pm$  0.019     &    5.533 $\pm$ 1.250   &    0.939   &   0.725   &   0.926   &   0.823   &  0.720   &   0.827         \\
%\bf Ours   &   \bf  4.171 $\pm$ 1.056  &      0.122 $\pm$ 0.027 &  \bf  0.036 $\pm$ 0.018 &  \bf  5.096 $\pm$ 1.003 &  0.942   & \bf 0.745  &  \bf  0.940  &    0.821   &  \bf 0.748  &  \bf 0.839 \\ 
\bf EyeFormer   &   \bf 4.069  $\pm$  1.089    &   0.122  $\pm$  0.029   &   \bf 0.036  $\pm$  0.018   &  \bf  5.043  $\pm$  1.052  &   0.942    & \bf  0.748    &  \bf  0.940   &  \bf  0.825   &   \bf 0.750   & \bf  0.841    \\
 \hline
\multicolumn{11}{c}{\textbf{Natural Scenes (OSIE dataset)}} \\
\hline
Itti--Koch &   3.180 $\pm$ 0.756 &   0.176 $\pm$ 0.039 &     0.061 $\pm$ 0.027 &  --- &   0.859 &  0.653 & 0.811 &  0.748 &  ---&  --- \\ 
SGC   &  2.992 $\pm$ 1.067  & 0.194 $\pm$ 0.071  &  0.073 $\pm$ 0.046 &  --- &  0.922 &  0.652 &  0.890 &  0.768 &  ---&  --- \\ 
Wang et al.  &  3.798  $\pm$ 1.128 & 0.227  $\pm$ 0.073 & 0.096  $\pm$ 0.060 &  --- &  0.886 & 0.641 &  0.841 &  0.700 &  ---&  --- \\ 
Le Meur et al.   &    3.027 $\pm$ 0.797 &   0.160 $\pm$ 0.476 &    0.057 $\pm$ 0.028 &  ---  & 0.892 &  0.653 & 0.865 & 0.770   & --- & ---   \\ 
STAR-FC   &  3.375 $\pm$ 1.300 &   0.228 $\pm$ 0.091 &   0.090 $\pm$ 0.067 &  ---  & 0.936 & 0.662 & 0.920 & 0.734 & --- & ---  \\ 
SaltiNet   &    3.439 $\pm$ 0.861 &  0.191 $\pm$ 0.052 &   0.065 $\pm$ 0.032 &   3.860 $\pm$ 0.814 &  0.895 & 0.641 & 0.872 & 0.719 &  0.573 & 0.740 \\ 
PathGAN    &  5.300 $\pm$ 1.197 &   0.323 $\pm$ 0.073 &   0.142 $\pm$ 0.085 & 5.454 $\pm$ 1.167 & 0.935 & 0.577 &  0.924 & 0.608 & 0.679 & 0.745 \\ 
IOR-ROI   &  2.495 $\pm$ 0.809 &   0.160 $\pm$ 0.055 &   0.060 $\pm$ 0.039 & 2.955 $\pm$ 0.768  & 0.914 &  \bf 0.704 & 0.889  & 0.812  & 0.629 & 0.790 \\ 
Chen et al.  &  \bf 2.183  $\pm$  0.949   &   0.125   $\pm$  0.056  &    0.045  $\pm$  0.028     &  2.636 $\pm$ 0.865   &   \bf 0.944   &   0.653   &   0.924   &   0.847   &  0.689   &   0.811         \\ 
\bf EyeFormer   &    2.193  $\pm$  0.831    &  \bf  0.115  $\pm$  0.042   &   \bf 0.044  $\pm$  0.026   &   \bf 2.562 $\pm$  0.756  &   \bf \bf 0.944    &   0.679    &  \bf  0.932   &  \bf  0.850   &  \bf 0.706   &   \bf 0.822    \\
\bottomrule
\end{tabular}
}

\caption{
    Quantitative scanpath evaluation,
    with the \textit{Mean} $\pm$ \textit{SD} reported for each metric.
    Our model outperforms the baseline models by most metrics on both natural
scenes and GUIs.
    ``Pretrained'' denotes testing via the pre-trained model. Others are trained with the same dataset.
    Boldface highlights the best result column-wise.
    Arrows indicate the importance relation's direction 
    (e.g., $\uparrow$ means ``higher is better'' ).
   Dashes (``-'') indicate methods unable to predict duration.
    % EyeFormer demonstrates superior performance across all metrics, 
    % apart from \textit{Shape} in \textit{MultiMatch}, indicating that our model produces more accurate results.
}
\Description{Quantitative scanpath evaluation comparing our model against other models, with the Mean and SD reported for each metric.}
\label{tbl:scanpaths_quantitative}
\end{table*}

\section{Experiments}

We show that our model has the unique capability of producing personalized predictions when given a few user scanpath samples. 
In addition, we present a comprehensive evaluation of the model against a number of recent models 
and across two very different types of stimuli: GUIs and natural scenes. 
Overall, we cover a large number of baselines and evaluation metrics for scanpath models.

\subsection{Datasets}

As mentioned before, we conducted experiments on both datasets of GUIs and natural scenes. 
Both datasets include multiple scanpaths from different viewers per image. 

\subsubsection{GUIs and Information Graphics}
We utilized the \textit{UEyes} dataset~\cite{jiang2023ueyes, jiang2023ueyes2}, 
including eye-tracking data (up to 7 seconds) from 62 participants on 1,980 images 
from four common types of GUIs and Information Graphics (posters, desktop GUIs, mobile GUIs, and webpages). 
This dataset was collected using an eye tracker in a laboratory setting, 
to guarantee precise fixation coordinates in the \emph{XY} plane, 
and the coordinate values were subjected to participant-specific calibration, 
accounting for relevant human factors such as eye--display distance~\cite{leiva2020understanding}.
We used the same training/test GUI image split as~\citet{jiang2023ueyes}: 
1,872 images in the training set and 108 images in the test set. 
The four GUI types are distributed evenly in each set.
In addition, we further established a training/test split for individual-level prediction: 
randomly assigning 53 viewers to the training set (85\%) and the remaining 9 to the test set (15\%). 
Our model was trained on the data collected from when the training viewers looked at the GUIs in the training images. 
In UEyes, most scanpaths have around 15 fixations (the average number of fixations per image is 15.3). 
We present more details of the dataset and implementation in the Supplementary Materials.

\subsubsection{Natural Scenes}
\Yue{We employed the \textit{OSIE} dataset~\cite{xu2014predicting}, a free-viewing dataset on natural scenes. 
This dataset comprises 700 images with 3-second eye movement data gathered from 15 participants, 
and has been widely used in previous research studies~\cite{chen2021predicting,sun2019visual}. 
We followed the same dataset split as in prior work (80\% training, 10\% validation, and 10\% testing data). 
We did not use datasets such as SALICON~\cite{jiang2015salicon} since they used mouse movements as a proxy for eye movements. 
Our model focuses on replicating actual scanpaths recorded by eye trackers.}

\subsection{Metrics}

We assessed the performance via commonly employed metrics for scanpath evaluation~\cite{anderson2015comparison, fahimi2021metrics}. 

    \subsubsection{Dynamic Time Warping (DTW)} DTW serves as a standard metric for similarity between two temporal sequences, 
    even when they have different lengths~\cite{berndt1994using, salvador2007toward}.
    It identifies the optimal match and calculates the distance between two scanpaths that preserves essential features.
    \subsubsection{Time Delay Embedding (TDE)} TDE focuses on assessing similarities between subscanpaths~\cite{tim1991embedology, wang2011simulating}. Thus, it offers a more nuanced evaluation compared to DTW, which focuses only on the overall comparison of entire scanpaths.
    \subsubsection{Eyenalysis} Eyenalysis finds the closest mapping between fixation points on the two scanpaths. 
    For each fixation point along one scanpath, it identifies the spatially closest fixation point on the other scanpath, and vice versa.~\cite{mathot2012eyenalysis}:
    Eyenalysis then measures the average distances for all the closest fixation pairs, 
    and thus focuses more on evaluating individual fixations instead of the sequences.

\subsubsection{Dynamic Time Warping with Duration (DTWD)} We extended DTW to capture duration; DTWD enables considering both fixation position and duration. 
we spatially align scanpaths using fixation positions, then compute DTWD value as 3D vectors $(x, y, t)$ of fixation position and duration.

\subsubsection{MultiMatch} MultiMatch~\cite{dewhurst2012depends} has five variants, each allowing us to assess important aspects of fixations in scanpaths: shape, direction, length, position, and duration. While DTWD evaluates spatial and temporal characteristics, MultiMatch excels in capturing additional features such as shape, direction, and length and gives an overall evaluation based on all these features.

\section{Results}

We show that our model EyeFormer  
1)~predicts individual-level scanpaths given a few user viewing samples;
2)~compares favorably against other models on population-level scanpath prediction;
3)~predicts both spatial and temporal characteristics of scanpaths on different types of stimuli including GUIs and natural scenes.

\subsection{Individual-Level Scanpath Prediction}

Prior research has not addressed the challenge of predicting personalized individual-level scanpaths. 
Full retraining for each new viewer, with more data, is impractical. 
Our model achieves a workable balance by generating scanpaths tailored to each viewer's viewing behaviors and idiosyncrasies while still permitting a single model's application for all viewers, without the burden of retraining.
We verified our model's ability to generate personalized scanpaths by proceeding from a few scanpath samples from the individual, 
thus confirming that the model can effectively capture each viewer's viewing preferences/behaviors and reflect them in its output.

When encountering a new viewer with a few samples available, 
the model updates the viewer embedding with $n_\mathrm{path}$ scanpaths obtained from that viewer (in our experiments, $n_\mathrm{path}=50$). 
We fine-tune the model by backpropagating from the scanpath samples 
so that it can predict scanpaths specific to this unique individual's viewing behaviors.

Since there is currently no established baseline method that can serve as a comparison point for our proposed approach, we compare the model personalized to the target test viewer against the model personalized to other test viewers to quantify the effectiveness in capturing the characteristics of individual viewers (\autoref{tbl:individual_table}). The results show that the errors of the model personalized on the target viewer are smaller than the models for other test viewers. Thus, the personalized model can better fit individual characteristics.
In addition, we present individual-level scanpath prediction qualitatively using a few examples in  \autoref{fig:shortened_fewshot_results}.
 We provide more results and explain the relationship between sample quantity and the performance in the Supplementary Materials.

\subsection{Population-Level Scanpath Prediction}

\Yue{
To assess how well our model predicts the spatiotemporal information of scanpaths, we compared its performance with prior scanpath models. 
We evaluate our model on both GUIs and natural scenes to show that it can be generalizable to different types of images.
For GUIs, we compared to Itti--Koch~\cite{itti1998model}, DeepGaze~III~\cite{kummerer2022deepgaze}, DeepGaze++~\cite{jiang2023ueyes}, SaltiNet~\cite{assens2017saltinet}, UMSS~\cite{wang2023scanpath}, PathGAN~\cite{Assens2018pathgan}, PathGAN++~\cite{jiang2023ueyes}, ScanGAN~\cite{martin2022scangan360}, ScanDMM~\cite{Sui_2023_CVPR}, and~\citet{chen2021predicting}.
For natural scenes, we compared to models focused on natural scenes: Itti--Koch~\cite{itti1998model}, SGC~\cite{sun2012what},~\citet{wang2011simulating}, Le Meur et al.~\cite{lemeur2015saccadic}, STAR-FC~\cite{wloka2018active}, SaltiNet~\cite{assens2017saltinet}, PathGAN~\cite{Assens2018pathgan}, IOR-ROI~\cite{sun2019visual},~\citet{chen2021predicting}. 
We trained these models with the same dataset split for a fair comparison.
We provide all individual scanpaths from all the viewers per training image, helping the models learn the underlying scanpath distribution.
Note that we did not combine the two datasets. We trained and analyzed them separately. All the baseline methods were also trained on each dataset separately.
}

\subsubsection{{Quantitative Evaluation}}

To account for variations in image sizes and minimize discrepancy-related errors, 
we normalized the fixation points' coordinates to the $[0,1]$ range. 
Specifically for training natural scenes, we used ResNet instead of ViT as the vision encoder to have a fair comparison to other baseline models, since prior work trained on the OSIE dataset~\cite{xu2014predicting} used ResNet as the encoder.
\autoref{tbl:scanpaths_quantitative} presents a comprehensive comparison encompassing these metrics. 
Our model outperforms the baseline models by most metrics on GUIs and natural scenes. 
It indicates that our model simulates scanpath trajectories more realistically. 
Among these models, only PathGAN, PathGAN++, SaltiNet, UMSS, IOR-ROI, and Chen et al. can predict temporal information. 
For example, the best baseline model Chen et al. predicts positions and duration separately. 
However, fixation positions and duration are highly correlated.
Our model excels by the DTWD and MultiMatch \texttt{Duration} metrics, 
attesting to its capability to yield more accurate results including the prediction of temporal information.

\subsubsection{{Qualitative Evaluation}}

Qualitative comparisons revealed that the predictions made by our model lie closer to the ground truth than the other models. \autoref{fig:shortened_scanpath_results} presents population-level prediction results showcasing the performance of our model. 
\autoref{fig:shortened_scanpath_qualitative} and \autoref{fig:scanpath_qualitative_chen} show the comparison of our model and the prior models. More results are in the Supplementary Materials.
While PathGAN++ and ScanGAN excel at generating realistic trajectories thanks to their discriminative component, the points they predict often fall outside the salient areas and tend to produce clusters.
In contrast, DeepGaze++ performs well in locating fixation points by applying post-processing to density maps. Nevertheless, it generates fixations in incorrect order; on account of the non-differentiable nature of the post-processing, the order is not optimized.
The Itti--Koch, SaltiNet, and UMSS generate scanpaths from saliency maps, encouraging fixations in salient areas, but they also do not optimize for the correct fixation order. 
Chen et al. tends to generate several clusters of closely grouped points since they improved the prediction of fixations while overlooking the need to spread consecutive points out more.
We additionally computed the clustering tendency error with the Laminarity metric~\cite{anderson2015comparison} (lower is better). The Laminarity value of our model is 73.137 and the value of Chen et al. is 178.072. Higher Laminarity indicates the model of Chen et al. has fixations clustering on locations where ground truth fixations are not clustered.
Finally, ScanDMM focuses relatively strongly on text elements. 
Our model assigns fixations to salient areas and attends to the points' order with greater precision.
It accomplishes this by using the salient-value reward ($r_\mathrm{sal}$) to emphasize points that lie within areas of interest and by employing the DTWD reward ($r_\mathrm{dtwd}$) to encourage more accurate trajectories.

\def\arraystretch{1}% 
\begin{table*}[t!]
\scalebox{0.85}{\setlength\tabcolsep{5pt}
\begin{tabular}{lcccccccccc}
\hline
\multirow{ 2}{*}{\textbf{Model}}  & \multirow{ 2}{*}{\textbf{DTW} $\downarrow$}  & \multirow{ 2}{*}{\textbf{TDE} $\downarrow$} & \multirow{ 2}{*}{\textbf{Eyenalysis} $\downarrow$} & \multirow{ 2}{*}{\textbf{DTWD} $\downarrow$} & \multicolumn{6}{|c}{\textbf{MultiMatch} $\uparrow$}  \\
 &  &  &   & &  \multicolumn{1}{|c}{\textbf{Shape}}   & \textbf{Direction}  &  \textbf{Length}   & \textbf{Position}   & \textbf{Duration}  & \textbf{Mean}  \\
\hline
\multicolumn{11}{c}{Population-Level Scanpath Prediction} \\
\hline
Ours w/o RLs  &    4.304 $\pm$ 1.309 &     0.143 $\pm$ 0.041 &   0.049 $\pm$ 0.024 &    5.299 $\pm$ 1.235 &  \bf 0.946  &  0.709  &  0.925  &  0.820  &  0.736  & 0.827 \\ 
Ours w/o $r_\mathrm{sal}$  &    4.099 $\pm$ 1.192  &     0.137 $\pm$ 0.036 &    0.045 $\pm$ 0.023 &    \bf 4.981 $\pm$ 1.131 &  \bf  0.946 & 
 0.713  &  0.928  &  0.825  &   \bf 0.752  &  0.833 \\
Ours w/o $r_\mathrm{dtw}$  &     5.277 $\pm$ 1.009  &     0.139 $\pm$ 0.025 &  \bf 0.036 $\pm$ 0.018 &    6.733 $\pm$ 1.038  &   0.913  &  0.736  &   0.907  &  0.789  &  0.673  & 0.804 \\
Ours w/o IOR  &  4.485  $\pm$ 1.353    &  0.177  $\pm$ 0.047  &   0.074  $\pm$ 0.034   &  5.327  $\pm$ 1.261       &   0.945  &    0.697   &   0.909    &    0.816     &     0.738  &  0.821  \\
% Ours   &     4.171 $\pm$ 1.056 &   \bf   0.122 $\pm$ 0.027 &  \bf  0.036 $\pm$ 0.018 &    5.096 $\pm$ 1.003 &  0.942   & \bf 0.745  &  \bf  0.940  &  \bf  0.821   &   0.748  &  \bf 0.839 \\ 
Ours   &   \bf 4.069  $\pm$  1.089    &  \bf 0.122  $\pm$  0.029   &   \bf 0.036  $\pm$  0.018   &    5.043  $\pm$  1.052  &   0.942    & \bf  0.748    &  \bf  0.940   &  \bf  0.825   &   \bf 0.750   & \bf  0.841    \\
\hline
\multicolumn{11}{c}{Individual-Level Scanpath Prediction for Training Viewers} \\
\hline
Ours w/o RLs &   \  4.362 $\pm$ 1.294 &      0.137 $\pm$ 0.039 &    0.046 $\pm$ 0.027 &    5.517 $\pm$ 1.200 & \bf  0.945  &   0.722 & 0.932 & 0.815   & 0.719  &  0.827   \\ 
Ours   &   \bf  4.164 $\pm$ 1.039 &   \bf   0.120 $\pm$ 0.027 &  \bf  0.035 $\pm$ 0.016 &   \bf 5.166 $\pm$ 0.998 &   0.937 & \bf  0.755  & \bf 0.936 &  \bf 0.824 &  \bf  0.738   &  \bf 0.838 \\ 
\bottomrule
\end{tabular}
}
\caption{
    Ablation study examining reinforcement learning's impact on population-level predictions and on individual-specific ones on training viewers on the UEyes dataset. The results highlight the importance of dynamic time warping and salient-value reward terms, 
    and also the use of inhibition of return.
}
\Description{Ablation study examining reinforcement learning's impact on population-level predictions and on individual-specific ones on training viewers on the UEyes dataset.}
\label{tbl:scanpaths_ablation}
\end{table*}

\subsection{Ablation Study}

\autoref{tbl:scanpaths_ablation} presents the results from an ablation study we performed 
for utilizing RL to produce both population- and individual-level scanpaths. 
The results reveal that a Transformer-only model does not produce satisfactory results
and incorporating RL greatly enhances the prediction of fixations and their duration.
With its population-level prediction, our RL model brings an improvement of 14.7\% and 26.5\% by the TDE and Eyenalysis metrics, respectively. 
This too is evidence that using RL increases the model's capacity to generate realistic fixations for scanpaths.
As for fixation duration, applying RL has a positive influence on prediction accuracy, 
demonstrated by the 4.8\% improvement shown by the DTWD metric.
Similar effects are visible with the individual-specific predictions connected with training users (i.e., the trained model's prediction of scanpaths for GUI images when given the IDs of particular training users). 
Additionally, the results highlight that the absence of either each type of reward or of inhibition of return leads to a decline in overall accuracy. We include more ablation studies in the Supplementary Materials.

\setlength{\tabcolsep}{0.9pt}
\def\arraystretch{0.2}% 
\begin{figure*}[!]
\def\w{0.250\linewidth}
 \centering
 \begin{tabular}{ccccc}
 &  Input Design   & \begin{tabular}{@{}c@{}}Population-Optimized Layout   \end{tabular}   & \begin{tabular}{@{}c@{}}Personalized Layout (Viewer 1) \end{tabular}  & \begin{tabular}{@{}c@{}}Personalized Layout (Viewer 2) \end{tabular}
\\
 \begin{turn}{90} 
\ \ \ \ \ \ \ \ \ \ \ \ \ \REVISION{Design 1}
\end{turn}  &
\includegraphics[width=\w]{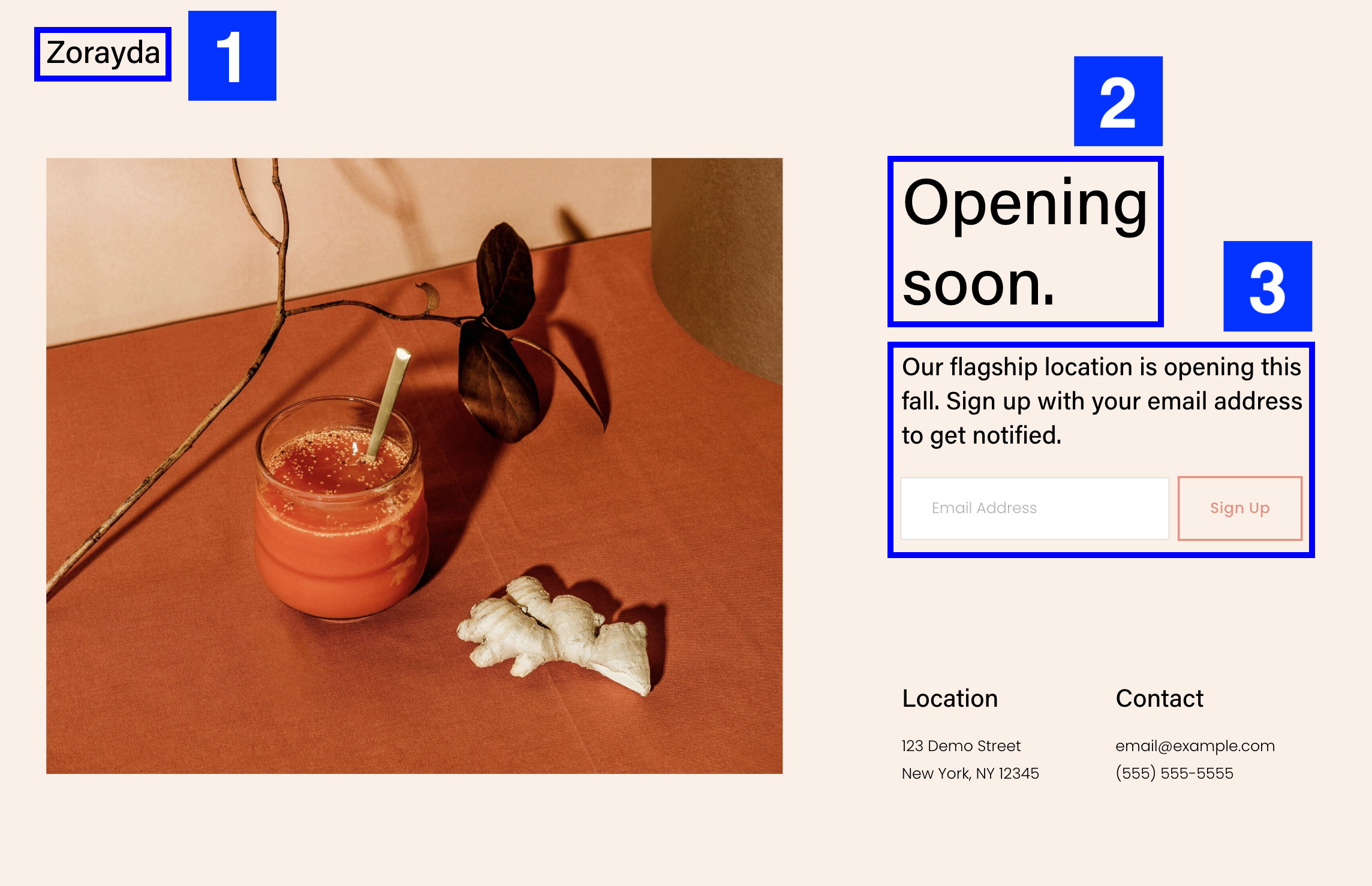} & 
\includegraphics[width=\w]{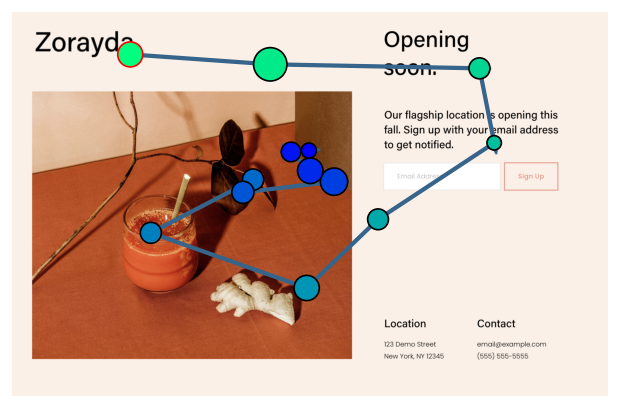} &
\includegraphics[width=\w]{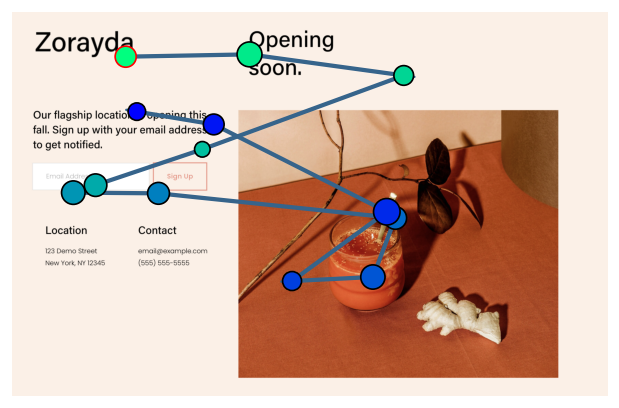} &
\includegraphics[width=\w]{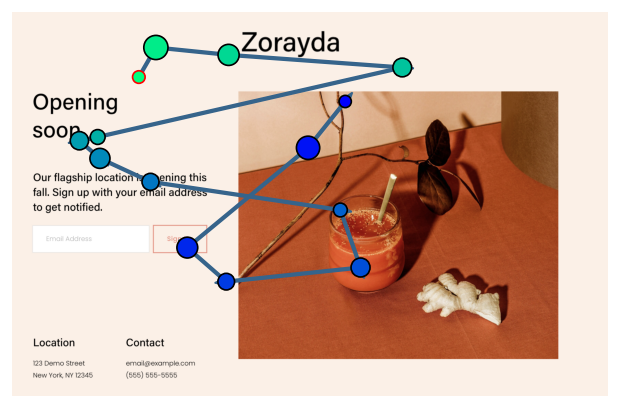} 
\\
& & Average Duration: 1.29 Seconds & Duration: 2.68 Seconds & Duration: 1.47 Seconds
\\
 \begin{turn}{90} 
\ \ \ \ \ \ \ \ \ \ \ \REVISION{Design 2}
\end{turn}  &
\includegraphics[width=\w]{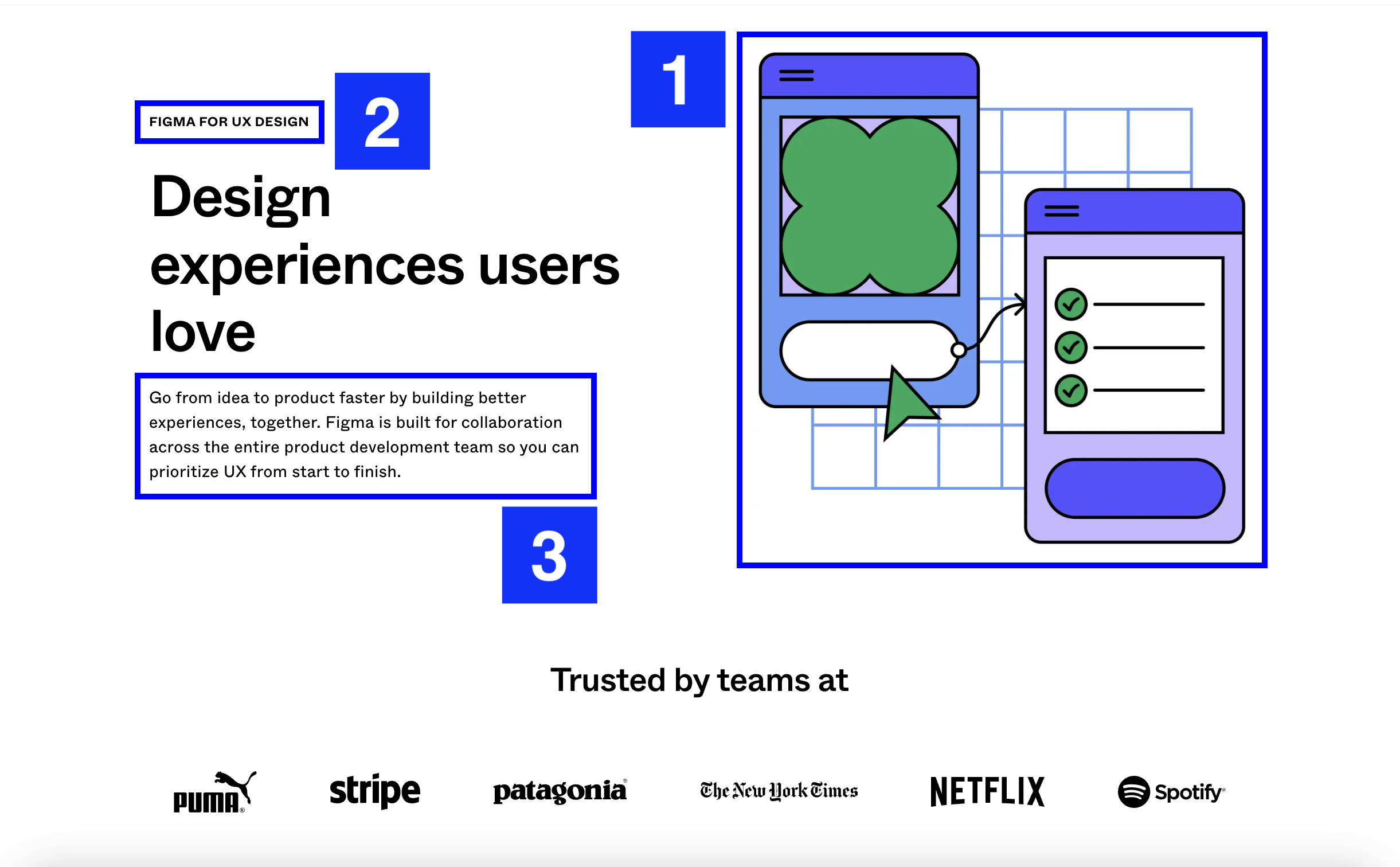} & 
\includegraphics[width=\w]{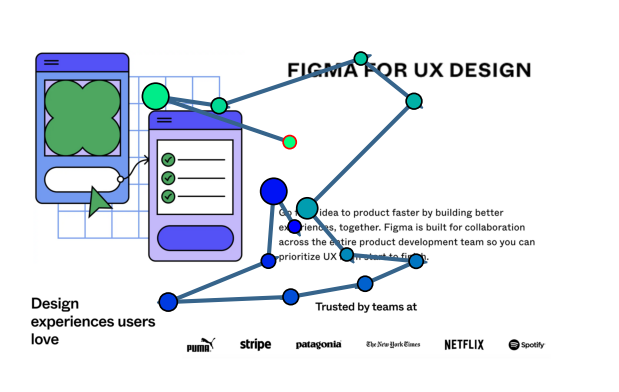} &
\includegraphics[width=\w]{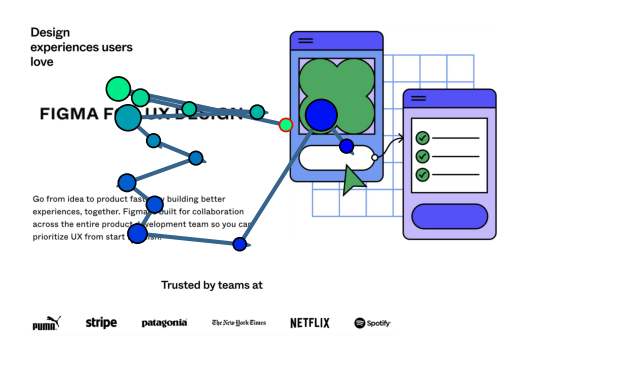} &
\includegraphics[width=\w]{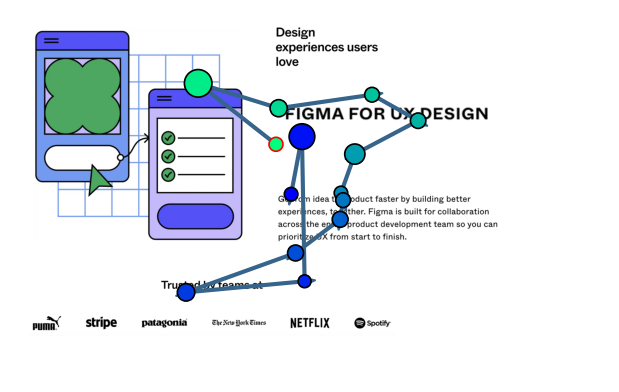}
\\
& & Average Duration: 2.75 Seconds & Duration: 4.06 Seconds & Duration: 2.91 Seconds
\\
\end{tabular}
\caption{
Given the input GUI design with the order of the three most important elements defined by the designer, we generate both the population-optimized layout and the personalized layout for each individual viewer. 
We present the average fixation duration of the selected important elements on the population-optimized layout across the tested viewers and the fixation duration on these elements for each shown personalized layout. Personalized layouts can attract more attention from their respective viewers to the target elements compared to the population-optimized layout.
}
\Description{This figure shows the results of application results. Given the input GUI design with the order of the three most important elements defined by the designer, we generate both the population-optimized layout and the personalized layout for each individual viewer.}
\label{fig:optimization}
\end{figure*}

\section{Application: Personalized Visual Flows}

EyeFormer enables the prediction of individual-level scanpaths, a capability we here apply to the problem of personalizing \emph{visual flows}. 
In model-assisted flow design, the designer picks GUI elements that should gain more attention than others \cite{fosco2020predicting}.
While previous work has demonstrated model-assisted personalization of graphical layouts \cite{todi2019individualising}, it has focused on visual search time and not visual flow. 
In our scenario, the designer gives a GUI layout and provides the order of the most important three or more elements that should be fixated first. Subsequently, our system outputs both population-optimized and individually-optimized layouts. The individually-optimized layouts are optimized based on the personalized scanpath prediction results. Specifically, given a viewer with $n_\mathrm{path}$ scanpath samples (in our experiments, $n_\mathrm{path}=50$), we generate the corresponding individually optimized layouts based on the predicted scanpaths at the individual level for this specific viewer.

\subsection{Formulation of Optimization Problem}

We define this problem as an optimization problem where we need to determine the positions and sizes of elements in a GUI based on the predicted personalized scanpaths.
To address this, we build on an integer programming-based layout optimizer~\cite{dayama2020grids}, 
which optimizes GUI layouts by considering the packing, alignment, and preferential positioning of GUI elements. 
We additionally introduce a constraint requiring adherence to the designer-specified fixation order, 
along with an objective score derived from the predictions of EyeFormer.

\subsubsection{Fixation Order Constraint} 
We denote the order of the three most important elements, $\text{elem}_1$, $\text{elem}_2$, and $\text{elem}_3$, which should be fixated first, as $[\text{elem}_1, \text{elem}_2, \text{elem}_3]$. It can be extended to handle more elements in a similar manner.
Firstly, for the predicted scanpath $[\hat{p}_1, \hat{p}_2, ..., \hat{p}_T]$, we identify the GUI element fixated on by each fixation point, denoted as $[\text{elem}_{\hat{p}_1}, \text{elem}_{\hat{p}_2}, ..., \text{elem}_{\hat{p}_T}]$.
Subsequently, we introduce a constraint to ensure that $[\text{elem}_1, \text{elem}_2, \text{elem}_3]$ is a sublist of the beginning of the deduplicated sequence of $[\text{elem}_{\hat{p}_1}, \text{elem}_{\hat{p}_2}, ..., \text{elem}_{\hat{p}_T}]$. In other words, the sequence $[\text{elem}_{\hat{p}_1}, \text{elem}_{\hat{p}_2}, ..., \text{elem}_{\hat{p}_T}]$ begins with repeated occurrences of $\text{elem}_1$, followed by $\text{elem}_2$, and subsequently by $\text{elem}_3$. This guarantees that the required fixation order specified by the designer is maintained.

\subsubsection{Fixation Duration Objective Term} 

We apply the fixation duration objective term for GUI layouts that satisfy the required order constraint to further determine the optimized layout. We denote the fixations corresponding to the subsequence of repeated occurrences of $\text{elem}_1$, followed by $\text{elem}_2$, and subsequently by $\text{elem}_3$ described above, as $[\hat{p}_1, \hat{p}_2, ..., \hat{p}_M]$, where $\hat{p}_M$ is the last fixation before moving attention to other elements. The objective is to select the layout that has the maximal sum of the duration of these elements $\sum_{m=1}^M t_{\hat{p}_m}$.

\subsection{Results} 

\autoref{fig:optimization} shows two exemplary results, and more results are given in Supplementary materials. Given an original GUI design, an ordered sequence of three most important GUI elements annotated, we generate both (i)~the population-optimized layout and (ii)~the personalized layout for each individual viewer. 
The population-optimized layout is based on the population-level scanpath prediction, which serves as the best compromise across viewers. 
The personalized layouts are based on personalized scanpath prediction so that each viewer fits the defined order and maximizes the duration fixated on the selected important elements. We test on 62 individual viewers. 
For ``Design 1'', 56 viewers would follow the desired viewing order of the annotated elements on the population-optimized layout, and the average duration is 1.29 seconds. 
All the viewers would follow the desired order on the corresponding personalized layout, and the average duration is 1.86 seconds, which is 44.19\% longer than on the population-optimized layout.
For ``Design 2'', 46 viewers would follow the desired viewing order of the annotated elements on the population-optimized layout, and the average duration is 2.75 seconds.
61 viewers would follow the desired order on the corresponding personalized layout, and the average duration is 3.19 seconds, which is 16\% longer than on the population-optimized layout.
The results show that personalized layouts can attract more attention from their respective viewers to the target elements compared to the population-optimized layout.

\section{Discussion}

EyeFormer is a Transformer-guided reinforcement learning method that establishes new state-of-the-art results. 
It is the first model that accurately models individual variability in scanpaths based on a few samples. 
In addition, it performs better or on par with previous state-of-the-art models on population-level prediction.
We found that the Transformer architecture is central for capturing long-range sequential dependencies from previous fixations with Gaussian distributions, but it is not enough to deliver optimal results. 
Combined with RL, it enables the generation of fixation sequences via optimization using non-differentiable objectives.

Using Transformers as the policy model in RL offers a natural representation 
for accurately capturing variability in scanning patterns across different stimuli and individuals. 
EyeFormer is able to cover both spatial and temporal characteristics of scanpaths across various kinds of stimuli and different individual viewing behaviors, which are important in understanding visual attention. We will release our code and model.

\paragraph{{Limitations and Future Work}}

EyeFormer opens the door to automated personalization of visual flows,
which enables GUI software to better respond to individuals' styles and expectations. By feeding some scanpath samples from a user, we demonstrated that personalized layouts can be generated for that user. Such capability can also be helpful for accessibility to optimize GUIs for people with viewing difficulties. It remains a practical challenge, however, how to obtain such data on an individual. We see at least a few options: 1)~using web cameras or other commodity devices with the user's permission, 2)~asking users to go through a calibration session where different designs are viewed, or 3)~inferring patterns via proxies like mouse movements. 

Our model is presently limited to fixed-length scanpaths 
since we concentrated on a limited time window of free-viewing behaviors,
based on UEyes's seven-second maximum length~\cite{jiang2023ueyes}, which allowed us to better compare against previous work. 
Future work should explore prediction that addresses varying lengths by predicting the final state.
In addition, this paper only focuses on fixation prediction. 
We acknowledge that viewing behaviors are far more complex, embracing other eye dynamics, {\em e.g.}, blinks, vestibulo-ocular reflexes, post-saccadic oscillations, 
which could be explored in future research.
Follow-up research could also investigate ways of reducing the number from the 50 scanpaths per viewer 
currently utilized to preclude suboptimal embeddings. 
Finally, the state-of-the-art scanpath-related metrics are designed primarily for natural scenes, 
so they may not fully capture the characteristics of scanpaths in GUI settings. 
Refining the metrics employed should afford a deeper understanding of these models' performance 
and, thereby, enhance the development of more effective methods.

% \appendix
% \section{Example galleries}

% In \autoref{fig:scanpath_qualitative} we compare and contrast different scanpath models,
% showing that ours is better aligned with ground-truth data.
% \autoref{fig:scanpath_results} provides further qualitative results of our model.

% \begin{acks}
% We thank Parvin Emami for providing feedback on an early version of this paper.
% This research is supported by [TODO: Grants/Funding IDs]
% \end{acks}

%\balance %% <-- Why this creates a blank page after the last figure?

\bibliographystyle{ACM-Reference-Format}
\bibliography{main}


\begin{thebibliography}{62}

%%% the Web address.

\ifx \showCODEN    \undefined \def \showCODEN     #1{\unskip}     \fi
\ifx \showDOI      \undefined \def \showDOI       #1{#1}\fi
\ifx \showISBNx    \undefined \def \showISBNx     #1{\unskip}     \fi
\ifx \showISBNxiii \undefined \def \showISBNxiii  #1{\unskip}     \fi
\ifx \showISSN     \undefined \def \showISSN      #1{\unskip}     \fi
\ifx \showLCCN     \undefined \def \showLCCN      #1{\unskip}     \fi
\ifx \shownote     \undefined \def \shownote      #1{#1}          \fi
\ifx \showarticletitle \undefined \def \showarticletitle #1{#1}   \fi
\ifx \showURL      \undefined \def \showURL       {\relax}        \fi
% invisible to TeX
\providecommand\bibfield[2]{#2}
\providecommand\bibinfo[2]{#2}
\providecommand\natexlab[1]{#1}
\providecommand\showeprint[2][]{arXiv:#2}

\bibitem[\protect\citeauthoryear{Anderson, Anderson, Kingstone, and Bischof}{Anderson et~al\mbox{.}}{2015}]%
        {anderson2015comparison}
\bibfield{author}{\bibinfo{person}{Nicola~C Anderson}, \bibinfo{person}{Fraser Anderson}, \bibinfo{person}{Alan Kingstone}, {and} \bibinfo{person}{Walter~F Bischof}.} \bibinfo{year}{2015}\natexlab{}.
\newblock \showarticletitle{A comparison of scanpath comparison methods}.
\newblock \bibinfo{journal}{\emph{Behavior research methods}} \bibinfo{volume}{47}, \bibinfo{number}{4} (\bibinfo{year}{2015}), \bibinfo{pages}{1377--1392}.
\newblock


\bibitem[\protect\citeauthoryear{Assens, Giro-i Nieto, McGuinness, and O’Connor}{Assens et~al\mbox{.}}{2017}]%
        {assens2017saltinet}
\bibfield{author}{\bibinfo{person}{Marc Assens}, \bibinfo{person}{Xavier Giro-i Nieto}, \bibinfo{person}{Kevin McGuinness}, {and} \bibinfo{person}{Noel~E. O’Connor}.} \bibinfo{year}{2017}\natexlab{}.
\newblock \showarticletitle{SaltiNet: Scan-Path Prediction on 360 Degree Images Using Saliency Volumes}. In \bibinfo{booktitle}{\emph{2017 IEEE International Conference on Computer Vision Workshops (ICCVW)}}. \bibinfo{pages}{2331--2338}.
\newblock
\urldef\tempurl%
\url{https://doi.org/10.1109/ICCVW.2017.275}
\showDOI{\tempurl}


\bibitem[\protect\citeauthoryear{Assens, i~Nieto, McGuinness, and O'Connor}{Assens et~al\mbox{.}}{2018}]%
        {Assens2018pathgan}
\bibfield{author}{\bibinfo{person}{Marc Assens}, \bibinfo{person}{Xavier~Giro i Nieto}, \bibinfo{person}{Kevin McGuinness}, {and} \bibinfo{person}{Noel~E. O'Connor}.} \bibinfo{year}{2018}\natexlab{}.
\newblock \showarticletitle{PathGAN: Visual Scanpath Prediction with Generative Adversarial Networks}.
\newblock \bibinfo{journal}{\emph{ECCV Workshop on Egocentric Perception, Interaction and Computing (EPIC)}}.
\newblock


\bibitem[\protect\citeauthoryear{Berndt and Clifford}{Berndt and Clifford}{1994}]%
        {berndt1994using}
\bibfield{author}{\bibinfo{person}{Donald~J Berndt} {and} \bibinfo{person}{James Clifford}.} \bibinfo{year}{1994}\natexlab{}.
\newblock \showarticletitle{Using dynamic time warping to find patterns in time series.}. In \bibinfo{booktitle}{\emph{KDD workshop}}, Vol.~\bibinfo{volume}{10}. Seattle, WA, USA:, \bibinfo{pages}{359--370}.
\newblock


\bibitem[\protect\citeauthoryear{Brohan, Gurney, and Dudek}{Brohan et~al\mbox{.}}{2010}]%
        {edc41706ff574381b7b93fe1403cd4ec}
\bibfield{author}{\bibinfo{person}{Kevin Brohan}, \bibinfo{person}{Kevin Gurney}, {and} \bibinfo{person}{Piotr Dudek}.} \bibinfo{year}{2010}\natexlab{}.
\newblock \showarticletitle{Using reinforcement learning to guide the development of self-organised feature maps for visual orienting}. In \bibinfo{booktitle}{\emph{Lecture Notes in Computer Science (including subseries Lecture Notes in Artificial Intelligence and Lecture Notes in Bioinformatics)|Lect. Notes Comput. Sci.}}, Vol.~\bibinfo{volume}{6353}. \bibinfo{publisher}{Springer Nature}, \bibinfo{address}{United States}, \bibinfo{pages}{180--189}.
\newblock
\showISBNx{3642158218}
\urldef\tempurl%
\url{https://doi.org/10.1007/978-3-642-15822-3_23}
\showDOI{\tempurl}
\newblock
\shownote{20th International Conference on Artificial Neural Networks, ICANN 2010 ; Conference date: 01-07-2010.}


\bibitem[\protect\citeauthoryear{Chen, Hsieh, and Gong}{Chen et~al\mbox{.}}{2021a}]%
        {ViT2021}
\bibfield{author}{\bibinfo{person}{Xiangning Chen}, \bibinfo{person}{Cho{-}Jui Hsieh}, {and} \bibinfo{person}{Boqing Gong}.} \bibinfo{year}{2021}\natexlab{a}.
\newblock \showarticletitle{When Vision Transformers Outperform ResNets without Pretraining or Strong Data Augmentations}.
\newblock \bibinfo{journal}{\emph{CoRR}}  \bibinfo{volume}{abs/2106.01548} (\bibinfo{year}{2021}).
\newblock
\showeprint[arXiv]{2106.01548}
\urldef\tempurl%
\url{https://arxiv.org/abs/2106.01548}
\showURL{%
\tempurl}


\bibitem[\protect\citeauthoryear{Chen, Jiang, and Zhao}{Chen et~al\mbox{.}}{2021b}]%
        {chen2021predicting}
\bibfield{author}{\bibinfo{person}{Xianyu Chen}, \bibinfo{person}{Ming Jiang}, {and} \bibinfo{person}{Qi Zhao}.} \bibinfo{year}{2021}\natexlab{b}.
\newblock \showarticletitle{Predicting human scanpaths in visual question answering}. In \bibinfo{booktitle}{\emph{Proceedings of the IEEE/CVF Conference on Computer Vision and Pattern Recognition}}. \bibinfo{pages}{10876--10885}.
\newblock


\bibitem[\protect\citeauthoryear{Chen and Sun}{Chen and Sun}{2018}]%
        {Chen2018_IJCAI}
\bibfield{author}{\bibinfo{person}{Zhenzhong Chen} {and} \bibinfo{person}{Wanjie Sun}.} \bibinfo{year}{2018}\natexlab{}.
\newblock \showarticletitle{Scanpath Prediction for Visual Attention Using IOR-ROI LSTM} \emph{(\bibinfo{series}{IJCAI'18})}. \bibinfo{publisher}{AAAI Press}, \bibinfo{pages}{642–648}.
\newblock
\showISBNx{9780999241127}


\bibitem[\protect\citeauthoryear{Dayama, Todi, Saarelainen, and Oulasvirta}{Dayama et~al\mbox{.}}{2020}]%
        {dayama2020grids}
\bibfield{author}{\bibinfo{person}{Niraj~Ramesh Dayama}, \bibinfo{person}{Kashyap Todi}, \bibinfo{person}{Taru Saarelainen}, {and} \bibinfo{person}{Antti Oulasvirta}.} \bibinfo{year}{2020}\natexlab{}.
\newblock \showarticletitle{Grids: Interactive layout design with integer programming}. In \bibinfo{booktitle}{\emph{Proceedings of the 2020 CHI Conference on Human Factors in Computing Systems}}. \bibinfo{pages}{1--13}.
\newblock


\bibitem[\protect\citeauthoryear{de~Belen, Bednarz, and Sowmya}{de~Belen et~al\mbox{.}}{2022}]%
        {de2022scanpathnet}
\bibfield{author}{\bibinfo{person}{Ryan Anthony~Jalova de Belen}, \bibinfo{person}{Tomasz Bednarz}, {and} \bibinfo{person}{Arcot Sowmya}.} \bibinfo{year}{2022}\natexlab{}.
\newblock \showarticletitle{ScanpathNet: A Recurrent Mixture Density Network for Scanpath Prediction}. In \bibinfo{booktitle}{\emph{2022 IEEE/CVF Conference on Computer Vision and Pattern Recognition Workshops (CVPRW)}}. IEEE, \bibinfo{pages}{5006--5016}.
\newblock


\bibitem[\protect\citeauthoryear{Dewhurst, Nystr{\"o}m, Jarodzka, Foulsham, Johansson, and Holmqvist}{Dewhurst et~al\mbox{.}}{2012}]%
        {dewhurst2012depends}
\bibfield{author}{\bibinfo{person}{Richard Dewhurst}, \bibinfo{person}{Marcus Nystr{\"o}m}, \bibinfo{person}{Halszka Jarodzka}, \bibinfo{person}{Tom Foulsham}, \bibinfo{person}{Roger Johansson}, {and} \bibinfo{person}{Kenneth Holmqvist}.} \bibinfo{year}{2012}\natexlab{}.
\newblock \showarticletitle{It depends on how you look at it: Scanpath comparison in multiple dimensions with MultiMatch, a vector-based approach}.
\newblock \bibinfo{journal}{\emph{Behavior research methods}}  \bibinfo{volume}{44} (\bibinfo{year}{2012}), \bibinfo{pages}{1079--1100}.
\newblock


\bibitem[\protect\citeauthoryear{Dosovitskiy, Beyer, Kolesnikov, Weissenborn, Zhai, Unterthiner, Dehghani, Minderer, Heigold, Gelly, Uszkoreit, and Houlsby}{Dosovitskiy et~al\mbox{.}}{2020}]%
        {Dosovitskiy2020ViT}
\bibfield{author}{\bibinfo{person}{Alexey Dosovitskiy}, \bibinfo{person}{Lucas Beyer}, \bibinfo{person}{Alexander Kolesnikov}, \bibinfo{person}{Dirk Weissenborn}, \bibinfo{person}{Xiaohua Zhai}, \bibinfo{person}{Thomas Unterthiner}, \bibinfo{person}{Mostafa Dehghani}, \bibinfo{person}{Matthias Minderer}, \bibinfo{person}{Georg Heigold}, \bibinfo{person}{Sylvain Gelly}, \bibinfo{person}{Jakob Uszkoreit}, {and} \bibinfo{person}{Neil Houlsby}.} \bibinfo{year}{2020}\natexlab{}.
\newblock \showarticletitle{An Image is Worth 16x16 Words: Transformers for Image Recognition at Scale}.
\newblock \bibinfo{journal}{\emph{CoRR}}  \bibinfo{volume}{abs/2010.11929} (\bibinfo{year}{2020}).
\newblock
\showeprint[arXiv]{2010.11929}
\urldef\tempurl%
\url{https://arxiv.org/abs/2010.11929}
\showURL{%
\tempurl}


\bibitem[\protect\citeauthoryear{Emami, Jiang, Guo, and Leiva}{Emami et~al\mbox{.}}{2024}]%
        {Emami24_etra}
\bibfield{author}{\bibinfo{person}{Parvin Emami}, \bibinfo{person}{Yue Jiang}, \bibinfo{person}{Zixin Guo}, {and} \bibinfo{person}{Luis~A. Leiva}.} \bibinfo{year}{2024}\natexlab{}.
\newblock \showarticletitle{Impact of Design Decisions in Scanpath Modeling}. In \bibinfo{booktitle}{\emph{Proceedings of the ACM Symposium on Eye Tracking Research \& Applications (ETRA)}}.
\newblock


\bibitem[\protect\citeauthoryear{Fahimi and Bruce}{Fahimi and Bruce}{2021}]%
        {fahimi2021metrics}
\bibfield{author}{\bibinfo{person}{Ramin Fahimi} {and} \bibinfo{person}{Neil~DB Bruce}.} \bibinfo{year}{2021}\natexlab{}.
\newblock \showarticletitle{On metrics for measuring scanpath similarity}.
\newblock \bibinfo{journal}{\emph{Behavior Research Methods}} \bibinfo{volume}{53}, \bibinfo{number}{2} (\bibinfo{year}{2021}), \bibinfo{pages}{609--628}.
\newblock


\bibitem[\protect\citeauthoryear{Fosco, Casser, Bedi, O'Donovan, Hertzmann, and Bylinskii}{Fosco et~al\mbox{.}}{2020}]%
        {fosco2020predicting}
\bibfield{author}{\bibinfo{person}{Camilo Fosco}, \bibinfo{person}{Vincent Casser}, \bibinfo{person}{Amish~Kumar Bedi}, \bibinfo{person}{Peter O'Donovan}, \bibinfo{person}{Aaron Hertzmann}, {and} \bibinfo{person}{Zoya Bylinskii}.} \bibinfo{year}{2020}\natexlab{}.
\newblock \showarticletitle{Predicting visual importance across graphic design types}. In \bibinfo{booktitle}{\emph{Proceedings of the 33rd Annual ACM Symposium on User Interface Software and Technology (UIST)}}. \bibinfo{pages}{249--260}.
\newblock


\bibitem[\protect\citeauthoryear{Itti, Koch, and Niebur}{Itti et~al\mbox{.}}{1998}]%
        {itti1998model}
\bibfield{author}{\bibinfo{person}{Laurent Itti}, \bibinfo{person}{Christof Koch}, {and} \bibinfo{person}{Ernst Niebur}.} \bibinfo{year}{1998}\natexlab{}.
\newblock \showarticletitle{A model of saliency-based visual attention for rapid scene analysis}.
\newblock \bibinfo{journal}{\emph{IEEE Transactions on pattern analysis and machine intelligence}} \bibinfo{volume}{20}, \bibinfo{number}{11} (\bibinfo{year}{1998}), \bibinfo{pages}{1254--1259}.
\newblock


\bibitem[\protect\citeauthoryear{Jiang, Huang, Duan, and Zhao}{Jiang et~al\mbox{.}}{2015}]%
        {jiang2015salicon}
\bibfield{author}{\bibinfo{person}{Ming Jiang}, \bibinfo{person}{Shengsheng Huang}, \bibinfo{person}{Juanyong Duan}, {and} \bibinfo{person}{Qi Zhao}.} \bibinfo{year}{2015}\natexlab{}.
\newblock \showarticletitle{SALICON: Saliency in Context}. In \bibinfo{booktitle}{\emph{2015 IEEE Conference on Computer Vision and Pattern Recognition (CVPR)}}. \bibinfo{pages}{1072--1080}.
\newblock
\urldef\tempurl%
\url{https://doi.org/10.1109/CVPR.2015.7298710}
\showDOI{\tempurl}


\bibitem[\protect\citeauthoryear{Jiang}{Jiang}{2024}]%
        {jiang2024computational2}
\bibfield{author}{\bibinfo{person}{Yue Jiang}.} \bibinfo{year}{2024}\natexlab{}.
\newblock \showarticletitle{Computational Representations for Graphical User Interfaces}. In \bibinfo{booktitle}{\emph{Extended Abstracts of the 2024 CHI Conference on Human Factors in Computing Systems}} \emph{(\bibinfo{series}{CHI EA '24})}.
\newblock


\bibitem[\protect\citeauthoryear{Jiang, Leiva, Houssel, Tavakoli, Kylmälä, and Oulasvirta}{Jiang et~al\mbox{.}}{2023a}]%
        {jiang2023ueyes}
\bibfield{author}{\bibinfo{person}{Yue Jiang}, \bibinfo{person}{Luis~A. Leiva}, \bibinfo{person}{Paul R.~B. Houssel}, \bibinfo{person}{Hamed~R. Tavakoli}, \bibinfo{person}{Julia Kylmälä}, {and} \bibinfo{person}{Antti Oulasvirta}.} \bibinfo{year}{2023}\natexlab{a}.
\newblock \showarticletitle{UEyes: Understanding Visual Saliency across User Interface Types}. In \bibinfo{booktitle}{\emph{Proceedings of the 2023 CHI Conference on Human Factors in Computing Systems}} \emph{(\bibinfo{series}{CHI '23})}.
\newblock


\bibitem[\protect\citeauthoryear{Jiang, Leiva, Rezazadegan~Tavakoli, RB~Houssel, Kylm{\"a}l{\"a}, and Oulasvirta}{Jiang et~al\mbox{.}}{2023b}]%
        {jiang2023ueyes2}
\bibfield{author}{\bibinfo{person}{Yue Jiang}, \bibinfo{person}{Luis~A Leiva}, \bibinfo{person}{Hamed Rezazadegan~Tavakoli}, \bibinfo{person}{Paul RB~Houssel}, \bibinfo{person}{Julia Kylm{\"a}l{\"a}}, {and} \bibinfo{person}{Antti Oulasvirta}.} \bibinfo{year}{2023}\natexlab{b}.
\newblock \showarticletitle{UEyes: An Eye-Tracking Dataset across User Interface Types}. In \bibinfo{booktitle}{\emph{Workshop Paper at the 2023 CHI Conference on Human Factors in Computing Systems}}.
\newblock


\bibitem[\protect\citeauthoryear{Jiang, Lu, Kliman-Silver, Lutteroth, Li, Nichols, and Stuerzlinger}{Jiang et~al\mbox{.}}{2024}]%
        {jiang2024computational}
\bibfield{author}{\bibinfo{person}{Yue Jiang}, \bibinfo{person}{Yuwen Lu}, \bibinfo{person}{Clara Kliman-Silver}, \bibinfo{person}{Christof Lutteroth}, \bibinfo{person}{Toby Jia-Jun Li}, \bibinfo{person}{Jeffrey Nichols}, {and} \bibinfo{person}{Wolfgang Stuerzlinger}.} \bibinfo{year}{2024}\natexlab{}.
\newblock \showarticletitle{Computational Methodologies for Understanding, Automating, and Evaluating User Interfaces}. In \bibinfo{booktitle}{\emph{Extended Abstracts of the 2024 CHI Conference on Human Factors in Computing Systems}} \emph{(\bibinfo{series}{CHI EA '24})}.
\newblock


\bibitem[\protect\citeauthoryear{Jiang, Lu, Lutteroth, Li, Nichols, and Stuerzlinger}{Jiang et~al\mbox{.}}{2023c}]%
        {jiang2023future}
\bibfield{author}{\bibinfo{person}{Yue Jiang}, \bibinfo{person}{Yuwen Lu}, \bibinfo{person}{Christof Lutteroth}, \bibinfo{person}{Toby Jia-Jun Li}, \bibinfo{person}{Jeffrey Nichols}, {and} \bibinfo{person}{Wolfgang Stuerzlinger}.} \bibinfo{year}{2023}\natexlab{c}.
\newblock \showarticletitle{The Future of Computational Approaches for Understanding and Adapting User Interfaces}. In \bibinfo{booktitle}{\emph{Extended Abstracts of the 2023 CHI Conference on Human Factors in Computing Systems}} (Hamburg, Germany) \emph{(\bibinfo{series}{CHI EA '23})}. \bibinfo{publisher}{Association for Computing Machinery}, \bibinfo{address}{New York, NY, USA}, Article \bibinfo{articleno}{367}, \bibinfo{numpages}{5}~pages.
\newblock
\showISBNx{9781450394222}
\urldef\tempurl%
\url{https://doi.org/10.1145/3544549.3573805}
\showDOI{\tempurl}


\bibitem[\protect\citeauthoryear{Jiang, Lu, Nichols, Stuerzlinger, Yu, Lutteroth, Li, Kumar, and Li}{Jiang et~al\mbox{.}}{2022}]%
        {jiang2022computational}
\bibfield{author}{\bibinfo{person}{Yue Jiang}, \bibinfo{person}{Yuwen Lu}, \bibinfo{person}{Jeffrey Nichols}, \bibinfo{person}{Wolfgang Stuerzlinger}, \bibinfo{person}{Chun Yu}, \bibinfo{person}{Christof Lutteroth}, \bibinfo{person}{Yang Li}, \bibinfo{person}{Ranjitha Kumar}, {and} \bibinfo{person}{Toby Jia-Jun Li}.} \bibinfo{year}{2022}\natexlab{}.
\newblock \showarticletitle{Computational Approaches for Understanding, Generating, and Adapting User Interfaces}. In \bibinfo{booktitle}{\emph{Extended Abstracts of the 2022 CHI Conference on Human Factors in Computing Systems}} (New Orleans, LA, USA) \emph{(\bibinfo{series}{CHI EA '22})}. \bibinfo{publisher}{Association for Computing Machinery}, \bibinfo{address}{New York, NY, USA}, Article \bibinfo{articleno}{74}, \bibinfo{numpages}{6}~pages.
\newblock
\showISBNx{9781450391566}
\urldef\tempurl%
\url{https://doi.org/10.1145/3491101.3504030}
\showDOI{\tempurl}


\bibitem[\protect\citeauthoryear{Klein, Klein, Lee, Cruickshanks, and Chappell}{Klein et~al\mbox{.}}{2001}]%
        {KLEIN20011757}
\bibfield{author}{\bibinfo{person}{Ronald Klein}, \bibinfo{person}{Barbara~E.K Klein}, \bibinfo{person}{Kristine~E Lee}, \bibinfo{person}{Karen~J Cruickshanks}, {and} \bibinfo{person}{Richard~J Chappell}.} \bibinfo{year}{2001}\natexlab{}.
\newblock \showarticletitle{Changes in visual acuity in a population over a 10-year period1 1Each author states that he/she has no proprietary interest in any aspect of this work.: The Beaver Dam eye study}.
\newblock \bibinfo{journal}{\emph{Ophthalmology}} \bibinfo{volume}{108}, \bibinfo{number}{10} (\bibinfo{year}{2001}), \bibinfo{pages}{1757--1766}.
\newblock
\showISSN{0161-6420}
\urldef\tempurl%
\url{https://doi.org/10.1016/S0161-6420(01)00769-2}
\showDOI{\tempurl}


\bibitem[\protect\citeauthoryear{K{\"u}mmerer, Bethge, and Wallis}{K{\"u}mmerer et~al\mbox{.}}{2022}]%
        {kummerer2022deepgaze}
\bibfield{author}{\bibinfo{person}{Matthias K{\"u}mmerer}, \bibinfo{person}{Matthias Bethge}, {and} \bibinfo{person}{Thomas~SA Wallis}.} \bibinfo{year}{2022}\natexlab{}.
\newblock \showarticletitle{DeepGaze III: Modeling free-viewing human scanpaths with deep learning}.
\newblock \bibinfo{journal}{\emph{Journal of Vision}} \bibinfo{volume}{22}, \bibinfo{number}{5} (\bibinfo{year}{2022}).
\newblock


\bibitem[\protect\citeauthoryear{{Le Meur} and Liu}{{Le Meur} and Liu}{2015}]%
        {lemeur2015saccadic}
\bibfield{author}{\bibinfo{person}{Olivier {Le Meur}} {and} \bibinfo{person}{Zhi Liu}.} \bibinfo{year}{2015}\natexlab{}.
\newblock \showarticletitle{Saccadic model of eye movements for free-viewing condition}.
\newblock \bibinfo{journal}{\emph{Vision Research}}  \bibinfo{volume}{116} (\bibinfo{year}{2015}), \bibinfo{pages}{152--164}.
\newblock
\showISSN{0042-6989}
\urldef\tempurl%
\url{https://doi.org/10.1016/j.visres.2014.12.026}
\showDOI{\tempurl}
\newblock
\shownote{Computational Models of Visual Attention.}


\bibitem[\protect\citeauthoryear{Leiva, Xue, Bansal, Tavakoli, K{\"o}ro{\dh}lu, Du, Dayama, and Oulasvirta}{Leiva et~al\mbox{.}}{2020}]%
        {leiva2020understanding}
\bibfield{author}{\bibinfo{person}{Luis~A Leiva}, \bibinfo{person}{Yunfei Xue}, \bibinfo{person}{Avya Bansal}, \bibinfo{person}{Hamed~R Tavakoli}, \bibinfo{person}{Tu{\dh}{\c{c}}e K{\"o}ro{\dh}lu}, \bibinfo{person}{Jingzhou Du}, \bibinfo{person}{Niraj~R Dayama}, {and} \bibinfo{person}{Antti Oulasvirta}.} \bibinfo{year}{2020}\natexlab{}.
\newblock \showarticletitle{Understanding visual saliency in mobile user interfaces}. In \bibinfo{booktitle}{\emph{Proceedings of the International conference on human-computer interaction with mobile devices and services}}. \bibinfo{pages}{1--12}.
\newblock


\bibitem[\protect\citeauthoryear{Liu, Zhang, Zhu, and Chang}{Liu et~al\mbox{.}}{2023}]%
        {Tieyuan2023}
\bibfield{author}{\bibinfo{person}{Tieyuan Liu}, \bibinfo{person}{Meng Zhang}, \bibinfo{person}{Chuangying Zhu}, {and} \bibinfo{person}{Liang Chang}.} \bibinfo{year}{2023}\natexlab{}.
\newblock \showarticletitle{Transformer-based convolutional forgetting knowledge tracking}.
\newblock \bibinfo{journal}{\emph{Scientific Reports}} (\bibinfo{year}{2023}).
\newblock


\bibitem[\protect\citeauthoryear{Martin, Gutierrez, and Masia}{Martin et~al\mbox{.}}{2022a}]%
        {martin2022probabilistic}
\bibfield{author}{\bibinfo{person}{Daniel Martin}, \bibinfo{person}{Diego Gutierrez}, {and} \bibinfo{person}{Belen Masia}.} \bibinfo{year}{2022}\natexlab{a}.
\newblock \showarticletitle{A probabilistic time-evolving approach to scanpath prediction}.
\newblock \bibinfo{journal}{\emph{arXiv preprint arXiv:2204.09404}} (\bibinfo{year}{2022}).
\newblock


\bibitem[\protect\citeauthoryear{Martin, Serrano, Bergman, Wetzstein, and Masia}{Martin et~al\mbox{.}}{2022b}]%
        {martin2022scangan360}
\bibfield{author}{\bibinfo{person}{Daniel Martin}, \bibinfo{person}{Ana Serrano}, \bibinfo{person}{Alexander~W Bergman}, \bibinfo{person}{Gordon Wetzstein}, {and} \bibinfo{person}{Belen Masia}.} \bibinfo{year}{2022}\natexlab{b}.
\newblock \showarticletitle{Scangan360: A generative model of realistic scanpaths for 360 images}.
\newblock \bibinfo{journal}{\emph{IEEE Transactions on Visualization and Computer Graphics}} \bibinfo{volume}{28}, \bibinfo{number}{5} (\bibinfo{year}{2022}), \bibinfo{pages}{2003--2013}.
\newblock


\bibitem[\protect\citeauthoryear{Mathot, Cristino, Gilchrist, and Theeuwes}{Mathot et~al\mbox{.}}{2012}]%
        {mathot2012eyenalysis}
\bibfield{author}{\bibinfo{person}{S Mathot}, \bibinfo{person}{F Cristino}, \bibinfo{person}{ID Gilchrist}, {and} \bibinfo{person}{J Theeuwes}.} \bibinfo{year}{2012}\natexlab{}.
\newblock \showarticletitle{Eyenalysis: A similarity measure for eye movement patterns}.
\newblock \bibinfo{journal}{\emph{Journal of Eye Movement Research}}  \bibinfo{volume}{5} (\bibinfo{year}{2012}), \bibinfo{pages}{1--15}.
\newblock


\bibitem[\protect\citeauthoryear{Minut and Mahadevan}{Minut and Mahadevan}{2001}]%
        {minut2001reinforcement}
\bibfield{author}{\bibinfo{person}{Silviu Minut} {and} \bibinfo{person}{Sridhar Mahadevan}.} \bibinfo{year}{2001}\natexlab{}.
\newblock \showarticletitle{A reinforcement learning model of selective visual attention}. In \bibinfo{booktitle}{\emph{Proceedings of the fifth international conference on Autonomous agents}}. \bibinfo{pages}{457--464}.
\newblock


\bibitem[\protect\citeauthoryear{Mondal, Yang, Ahn, Samaras, Zelinsky, and Hoai}{Mondal et~al\mbox{.}}{2023}]%
        {Mondal_2023_CVPR}
\bibfield{author}{\bibinfo{person}{Sounak Mondal}, \bibinfo{person}{Zhibo Yang}, \bibinfo{person}{Seoyoung Ahn}, \bibinfo{person}{Dimitris Samaras}, \bibinfo{person}{Gregory Zelinsky}, {and} \bibinfo{person}{Minh Hoai}.} \bibinfo{year}{2023}\natexlab{}.
\newblock \showarticletitle{Gazeformer: Scalable, Effective and Fast Prediction of Goal-Directed Human Attention}. In \bibinfo{booktitle}{\emph{Proceedings of the IEEE/CVF Conference on Computer Vision and Pattern Recognition (CVPR)}}. \bibinfo{pages}{1441--1450}.
\newblock


\bibitem[\protect\citeauthoryear{Mousavi, Schukat, Howley, Borji, and Mozayani}{Mousavi et~al\mbox{.}}{2017}]%
        {mousavi2017learning}
\bibfield{author}{\bibinfo{person}{Sajad Mousavi}, \bibinfo{person}{Michael Schukat}, \bibinfo{person}{Enda Howley}, \bibinfo{person}{Ali Borji}, {and} \bibinfo{person}{Nasser Mozayani}.} \bibinfo{year}{2017}\natexlab{}.
\newblock \bibinfo{title}{Learning to predict where to look in interactive environments using deep recurrent q-learning}.
\newblock
\newblock
\showeprint[arxiv]{1612.05753}~[cs.CV]


\bibitem[\protect\citeauthoryear{Ognibene, Balkenius, and Baldassarre}{Ognibene et~al\mbox{.}}{2008}]%
        {ognibene2008reinforcement}
\bibfield{author}{\bibinfo{person}{Dimitri Ognibene}, \bibinfo{person}{Christian Balkenius}, {and} \bibinfo{person}{Gianluca Baldassarre}.} \bibinfo{year}{2008}\natexlab{}.
\newblock \showarticletitle{A reinforcement-learning model of top-down attention based on a potential-action map}.
\newblock In \bibinfo{booktitle}{\emph{The Challenge of Anticipation: A Unifying Framework for the Analysis and Design of Artificial Cognitive Systems}}. \bibinfo{publisher}{Springer}, \bibinfo{pages}{161--184}.
\newblock


\bibitem[\protect\citeauthoryear{Pang, Cao, Lau, and Chan}{Pang et~al\mbox{.}}{2016}]%
        {pang2016directing}
\bibfield{author}{\bibinfo{person}{Xufang Pang}, \bibinfo{person}{Ying Cao}, \bibinfo{person}{Rynson~WH Lau}, {and} \bibinfo{person}{Antoni~B Chan}.} \bibinfo{year}{2016}\natexlab{}.
\newblock \showarticletitle{Directing user attention via visual flow on web designs}.
\newblock \bibinfo{journal}{\emph{ACM Transactions on Graphics (TOG)}} \bibinfo{volume}{35}, \bibinfo{number}{6} (\bibinfo{year}{2016}), \bibinfo{pages}{1--11}.
\newblock


\bibitem[\protect\citeauthoryear{Qiu, Rong, Liang, and Tu}{Qiu et~al\mbox{.}}{2023}]%
        {qiu2023visual}
\bibfield{author}{\bibinfo{person}{Mengyu Qiu}, \bibinfo{person}{Quan Rong}, \bibinfo{person}{Dong Liang}, {and} \bibinfo{person}{Huawei Tu}.} \bibinfo{year}{2023}\natexlab{}.
\newblock \showarticletitle{Visual ScanPath Transformer: Guiding Computers to See the World}. In \bibinfo{booktitle}{\emph{2023 IEEE International Symposium on Mixed and Augmented Reality (ISMAR)}}. IEEE, \bibinfo{pages}{223--232}.
\newblock


\bibitem[\protect\citeauthoryear{Ranzato, Chopra, Auli, and Zaremba}{Ranzato et~al\mbox{.}}{2015}]%
        {ranzato2015sequence}
\bibfield{author}{\bibinfo{person}{Marc'Aurelio Ranzato}, \bibinfo{person}{Sumit Chopra}, \bibinfo{person}{Michael Auli}, {and} \bibinfo{person}{Wojciech Zaremba}.} \bibinfo{year}{2015}\natexlab{}.
\newblock \showarticletitle{Sequence level training with recurrent neural networks}.
\newblock \bibinfo{journal}{\emph{arXiv preprint arXiv:1511.06732}} (\bibinfo{year}{2015}).
\newblock


\bibitem[\protect\citeauthoryear{Rennie, Marcheret, Mroueh, Ross, and Goel}{Rennie et~al\mbox{.}}{2017}]%
        {rennie2017self}
\bibfield{author}{\bibinfo{person}{Steven~J Rennie}, \bibinfo{person}{Etienne Marcheret}, \bibinfo{person}{Youssef Mroueh}, \bibinfo{person}{Jerret Ross}, {and} \bibinfo{person}{Vaibhava Goel}.} \bibinfo{year}{2017}\natexlab{}.
\newblock \showarticletitle{Self-critical sequence training for image captioning}. In \bibinfo{booktitle}{\emph{Proceedings of the IEEE conference on computer vision and pattern recognition}}. \bibinfo{pages}{7008--7024}.
\newblock


\bibitem[\protect\citeauthoryear{{Rezazadegan Tavakoli}, Rahtu, and Heikkilä}{{Rezazadegan Tavakoli} et~al\mbox{.}}{2013}]%
        {REZAZADEGANTAVAKOLI2013686}
\bibfield{author}{\bibinfo{person}{Hamed {Rezazadegan Tavakoli}}, \bibinfo{person}{Esa Rahtu}, {and} \bibinfo{person}{Janne Heikkilä}.} \bibinfo{year}{2013}\natexlab{}.
\newblock \showarticletitle{Stochastic bottom–up fixation prediction and saccade generation}.
\newblock \bibinfo{journal}{\emph{Image and Vision Computing}} \bibinfo{volume}{31}, \bibinfo{number}{9} (\bibinfo{year}{2013}), \bibinfo{pages}{686--693}.
\newblock
\showISSN{0262-8856}
\urldef\tempurl%
\url{https://doi.org/10.1016/j.imavis.2013.06.006}
\showDOI{\tempurl}


\bibitem[\protect\citeauthoryear{Rosenholtz, Dorai, and Freeman}{Rosenholtz et~al\mbox{.}}{2011}]%
        {Rosenholtz11}
\bibfield{author}{\bibinfo{person}{Ruth Rosenholtz}, \bibinfo{person}{Amal Dorai}, {and} \bibinfo{person}{Rosalind Freeman}.} \bibinfo{year}{2011}\natexlab{}.
\newblock \showarticletitle{Do Predictions of Visual Perception Aid Design?}
\newblock \bibinfo{journal}{\emph{ACM Trans. Appl. Percept.}} \bibinfo{volume}{8}, \bibinfo{number}{2} (\bibinfo{year}{2011}).
\newblock


\bibitem[\protect\citeauthoryear{Salvador and Chan}{Salvador and Chan}{2007}]%
        {salvador2007toward}
\bibfield{author}{\bibinfo{person}{Stan Salvador} {and} \bibinfo{person}{Philip Chan}.} \bibinfo{year}{2007}\natexlab{}.
\newblock \showarticletitle{Toward accurate dynamic time warping in linear time and space}.
\newblock \bibinfo{journal}{\emph{Intelligent Data Analysis}} \bibinfo{volume}{11}, \bibinfo{number}{5} (\bibinfo{year}{2007}), \bibinfo{pages}{561--580}.
\newblock


\bibitem[\protect\citeauthoryear{Schwinn, Precup, Eskofier, and Zanca}{Schwinn et~al\mbox{.}}{2022}]%
        {schwinn2022behind}
\bibfield{author}{\bibinfo{person}{Leo Schwinn}, \bibinfo{person}{Doina Precup}, \bibinfo{person}{Bj{\"o}rn Eskofier}, {and} \bibinfo{person}{Dario Zanca}.} \bibinfo{year}{2022}\natexlab{}.
\newblock \showarticletitle{Behind the Machine's Gaze: Neural Networks with Biologically-inspired Constraints Exhibit Human-like Visual Attention}.
\newblock \bibinfo{journal}{\emph{arXiv preprint arXiv:2204.09093}} (\bibinfo{year}{2022}).
\newblock


\bibitem[\protect\citeauthoryear{Still and Masciocchi}{Still and Masciocchi}{2010}]%
        {Still10}
\bibfield{author}{\bibinfo{person}{Jeremiah~D. Still} {and} \bibinfo{person}{Christopher~M. Masciocchi}.} \bibinfo{year}{2010}\natexlab{}.
\newblock \showarticletitle{A Saliency Model Predicts Fixations in Web Interfaces}. In \bibinfo{booktitle}{\emph{Proc. MDDAUI Workshop}}.
\newblock


\bibitem[\protect\citeauthoryear{Sui, Fang, Zhu, Wang, and Wang}{Sui et~al\mbox{.}}{2023}]%
        {Sui_2023_CVPR}
\bibfield{author}{\bibinfo{person}{Xiangjie Sui}, \bibinfo{person}{Yuming Fang}, \bibinfo{person}{Hanwei Zhu}, \bibinfo{person}{Shiqi Wang}, {and} \bibinfo{person}{Zhou Wang}.} \bibinfo{year}{2023}\natexlab{}.
\newblock \showarticletitle{ScanDMM: A Deep Markov Model of Scanpath Prediction for 360deg Images}. In \bibinfo{booktitle}{\emph{Proceedings of the IEEE/CVF Conference on Computer Vision and Pattern Recognition (CVPR)}}. \bibinfo{pages}{6989--6999}.
\newblock


\bibitem[\protect\citeauthoryear{Sun, Chen, and Wu}{Sun et~al\mbox{.}}{2019}]%
        {sun2019visual}
\bibfield{author}{\bibinfo{person}{Wanjie Sun}, \bibinfo{person}{Zhenzhong Chen}, {and} \bibinfo{person}{Feng Wu}.} \bibinfo{year}{2019}\natexlab{}.
\newblock \showarticletitle{Visual scanpath prediction using IOR-ROI recurrent mixture density network}.
\newblock \bibinfo{journal}{\emph{IEEE transactions on pattern analysis and machine intelligence}} \bibinfo{volume}{43}, \bibinfo{number}{6} (\bibinfo{year}{2019}), \bibinfo{pages}{2101--2118}.
\newblock


\bibitem[\protect\citeauthoryear{Sun, Yao, and Ji}{Sun et~al\mbox{.}}{2012}]%
        {sun2012what}
\bibfield{author}{\bibinfo{person}{Xiaoshuai Sun}, \bibinfo{person}{Hongxun Yao}, {and} \bibinfo{person}{Rongrong Ji}.} \bibinfo{year}{2012}\natexlab{}.
\newblock \showarticletitle{What are we looking for: Towards statistical modeling of saccadic eye movements and visual saliency}. In \bibinfo{booktitle}{\emph{2012 IEEE Conference on Computer Vision and Pattern Recognition}}. \bibinfo{pages}{1552--1559}.
\newblock
\urldef\tempurl%
\url{https://doi.org/10.1109/CVPR.2012.6247846}
\showDOI{\tempurl}


\bibitem[\protect\citeauthoryear{Sutton and Barto}{Sutton and Barto}{2018}]%
        {sutton2018reinforcement}
\bibfield{author}{\bibinfo{person}{Richard~S Sutton} {and} \bibinfo{person}{Andrew~G Barto}.} \bibinfo{year}{2018}\natexlab{}.
\newblock \bibinfo{booktitle}{\emph{Reinforcement learning: An introduction}}.
\newblock \bibinfo{publisher}{MIT press}.
\newblock


\bibitem[\protect\citeauthoryear{Tang and Agrawal}{Tang and Agrawal}{2020}]%
        {tang2020discretizing}
\bibfield{author}{\bibinfo{person}{Yunhao Tang} {and} \bibinfo{person}{Shipra Agrawal}.} \bibinfo{year}{2020}\natexlab{}.
\newblock \showarticletitle{Discretizing continuous action space for on-policy optimization}. In \bibinfo{booktitle}{\emph{Proceedings of the aaai conference on artificial intelligence}}, Vol.~\bibinfo{volume}{34}. \bibinfo{pages}{5981--5988}.
\newblock


\bibitem[\protect\citeauthoryear{Tim, James, and Martin}{Tim et~al\mbox{.}}{1991}]%
        {tim1991embedology}
\bibfield{author}{\bibinfo{person}{Sauer Tim}, \bibinfo{person}{A~Yorke James}, {and} \bibinfo{person}{Casdagli Martin}.} \bibinfo{year}{1991}\natexlab{}.
\newblock \showarticletitle{Embedology}.
\newblock \bibinfo{journal}{\emph{Journal of statistical Physics}} \bibinfo{volume}{65}, \bibinfo{number}{3-4} (\bibinfo{year}{1991}), \bibinfo{pages}{579--616}.
\newblock


\bibitem[\protect\citeauthoryear{Todi, Jokinen, Luyten, and Oulasvirta}{Todi et~al\mbox{.}}{2019}]%
        {todi2019individualising}
\bibfield{author}{\bibinfo{person}{Kashyap Todi}, \bibinfo{person}{Jussi Jokinen}, \bibinfo{person}{Kris Luyten}, {and} \bibinfo{person}{Antti Oulasvirta}.} \bibinfo{year}{2019}\natexlab{}.
\newblock \showarticletitle{Individualising graphical layouts with predictive visual search models}.
\newblock \bibinfo{journal}{\emph{ACM Transactions on Interactive Intelligent Systems (TiiS)}} \bibinfo{volume}{10}, \bibinfo{number}{1} (\bibinfo{year}{2019}), \bibinfo{pages}{1--24}.
\newblock


\bibitem[\protect\citeauthoryear{Vaswani, Shazeer, Parmar, Uszkoreit, Jones, Gomez, Kaiser, and Polosukhin}{Vaswani et~al\mbox{.}}{2017}]%
        {vaswani2017attention}
\bibfield{author}{\bibinfo{person}{Ashish Vaswani}, \bibinfo{person}{Noam Shazeer}, \bibinfo{person}{Niki Parmar}, \bibinfo{person}{Jakob Uszkoreit}, \bibinfo{person}{Llion Jones}, \bibinfo{person}{Aidan~N Gomez}, \bibinfo{person}{{\L}ukasz Kaiser}, {and} \bibinfo{person}{Illia Polosukhin}.} \bibinfo{year}{2017}\natexlab{}.
\newblock \showarticletitle{Attention is all you need}.
\newblock \bibinfo{journal}{\emph{Advances in neural information processing systems}}  \bibinfo{volume}{30} (\bibinfo{year}{2017}).
\newblock


\bibitem[\protect\citeauthoryear{Verma and Sen}{Verma and Sen}{2019}]%
        {Verma2019}
\bibfield{author}{\bibinfo{person}{Ashish Verma} {and} \bibinfo{person}{Debashis Sen}.} \bibinfo{year}{2019}\natexlab{}.
\newblock \showarticletitle{HMM-based Convolutional LSTM for Visual Scanpath Prediction}. In \bibinfo{booktitle}{\emph{2019 27th European Signal Processing Conference (EUSIPCO)}}. \bibinfo{pages}{1--5}.
\newblock
\urldef\tempurl%
\url{https://doi.org/10.23919/EUSIPCO.2019.8902643}
\showDOI{\tempurl}


\bibitem[\protect\citeauthoryear{Wang, Li, and Smola}{Wang et~al\mbox{.}}{2019}]%
        {Wang2019}
\bibfield{author}{\bibinfo{person}{Chenguang Wang}, \bibinfo{person}{Mu Li}, {and} \bibinfo{person}{Alexander~J. Smola}.} \bibinfo{year}{2019}\natexlab{}.
\newblock \showarticletitle{Language Models with Transformers}.
\newblock \bibinfo{journal}{\emph{CoRR}}  \bibinfo{volume}{abs/1904.09408} (\bibinfo{year}{2019}).
\newblock
\showeprint[arXiv]{1904.09408}
\urldef\tempurl%
\url{http://arxiv.org/abs/1904.09408}
\showURL{%
\tempurl}


\bibitem[\protect\citeauthoryear{Wang, Chen, Wang, Jiang, Fang, and Yao}{Wang et~al\mbox{.}}{2011}]%
        {wang2011simulating}
\bibfield{author}{\bibinfo{person}{Wei Wang}, \bibinfo{person}{Cheng Chen}, \bibinfo{person}{Yizhou Wang}, \bibinfo{person}{Tingting Jiang}, \bibinfo{person}{Fang Fang}, {and} \bibinfo{person}{Yuan Yao}.} \bibinfo{year}{2011}\natexlab{}.
\newblock \showarticletitle{Simulating human saccadic scanpaths on natural images}. In \bibinfo{booktitle}{\emph{CVPR 2011}}. \bibinfo{pages}{441--448}.
\newblock
\urldef\tempurl%
\url{https://doi.org/10.1109/CVPR.2011.5995423}
\showDOI{\tempurl}


\bibitem[\protect\citeauthoryear{Wang, Bulling, et~al\mbox{.}}{Wang et~al\mbox{.}}{2023}]%
        {wang2023scanpath}
\bibfield{author}{\bibinfo{person}{Yao Wang}, \bibinfo{person}{Andreas Bulling}, {et~al\mbox{.}}} \bibinfo{year}{2023}\natexlab{}.
\newblock \showarticletitle{Scanpath prediction on information visualisations}.
\newblock \bibinfo{journal}{\emph{IEEE Transactions on Visualization and Computer Graphics}} (\bibinfo{year}{2023}).
\newblock


\bibitem[\protect\citeauthoryear{Williams}{Williams}{1992}]%
        {williams1992simple}
\bibfield{author}{\bibinfo{person}{Ronald~J Williams}.} \bibinfo{year}{1992}\natexlab{}.
\newblock \showarticletitle{Simple statistical gradient-following algorithms for connectionist reinforcement learning}.
\newblock \bibinfo{journal}{\emph{Reinforcement learning}} (\bibinfo{year}{1992}), \bibinfo{pages}{5--32}.
\newblock


\bibitem[\protect\citeauthoryear{Wloka, Kotseruba, and Tsotsos}{Wloka et~al\mbox{.}}{2018}]%
        {wloka2018active}
\bibfield{author}{\bibinfo{person}{Calden Wloka}, \bibinfo{person}{Iuliia Kotseruba}, {and} \bibinfo{person}{John~K. Tsotsos}.} \bibinfo{year}{2018}\natexlab{}.
\newblock \showarticletitle{Active Fixation Control to Predict Saccade Sequences}. In \bibinfo{booktitle}{\emph{2018 IEEE/CVF Conference on Computer Vision and Pattern Recognition}}. \bibinfo{pages}{3184--3193}.
\newblock
\urldef\tempurl%
\url{https://doi.org/10.1109/CVPR.2018.00336}
\showDOI{\tempurl}


\bibitem[\protect\citeauthoryear{Xia, Han, Qi, and Shi}{Xia et~al\mbox{.}}{2019}]%
        {Xia2019}
\bibfield{author}{\bibinfo{person}{Chen Xia}, \bibinfo{person}{Junwei Han}, \bibinfo{person}{Fei Qi}, {and} \bibinfo{person}{Guangming Shi}.} \bibinfo{year}{2019}\natexlab{}.
\newblock \showarticletitle{Predicting Human Saccadic Scanpaths Based on Iterative Representation Learning}.
\newblock \bibinfo{journal}{\emph{IEEE Transactions on Image Processing}} \bibinfo{volume}{28}, \bibinfo{number}{7} (\bibinfo{year}{2019}), \bibinfo{pages}{3502--3515}.
\newblock
\urldef\tempurl%
\url{https://doi.org/10.1109/TIP.2019.2897966}
\showDOI{\tempurl}


\bibitem[\protect\citeauthoryear{Xu, Jiang, Wang, Kankanhalli, and Zhao}{Xu et~al\mbox{.}}{2014}]%
        {xu2014predicting}
\bibfield{author}{\bibinfo{person}{Juan Xu}, \bibinfo{person}{Ming Jiang}, \bibinfo{person}{Shuo Wang}, \bibinfo{person}{Mohan~S Kankanhalli}, {and} \bibinfo{person}{Qi Zhao}.} \bibinfo{year}{2014}\natexlab{}.
\newblock \showarticletitle{Predicting human gaze beyond pixels}.
\newblock \bibinfo{journal}{\emph{Journal of vision}} \bibinfo{volume}{14}, \bibinfo{number}{1} (\bibinfo{year}{2014}), \bibinfo{pages}{28--28}.
\newblock


\bibitem[\protect\citeauthoryear{Xu, Song, Wang, Qiao, Huo, and Wang}{Xu et~al\mbox{.}}{2018}]%
        {xu2018predicting}
\bibfield{author}{\bibinfo{person}{Mai Xu}, \bibinfo{person}{Yuhang Song}, \bibinfo{person}{Jianyi Wang}, \bibinfo{person}{MingLang Qiao}, \bibinfo{person}{Liangyu Huo}, {and} \bibinfo{person}{Zulin Wang}.} \bibinfo{year}{2018}\natexlab{}.
\newblock \showarticletitle{Predicting head movement in panoramic video: A deep reinforcement learning approach}.
\newblock \bibinfo{journal}{\emph{IEEE transactions on pattern analysis and machine intelligence}} \bibinfo{volume}{41}, \bibinfo{number}{11} (\bibinfo{year}{2018}), \bibinfo{pages}{2693--2708}.
\newblock


\bibitem[\protect\citeauthoryear{Yang, Huang, Chen, Wei, Ahn, Zelinsky, Samaras, and Hoai}{Yang et~al\mbox{.}}{2020}]%
        {yang2020predicting}
\bibfield{author}{\bibinfo{person}{Zhibo Yang}, \bibinfo{person}{Lihan Huang}, \bibinfo{person}{Yupei Chen}, \bibinfo{person}{Zijun Wei}, \bibinfo{person}{Seoyoung Ahn}, \bibinfo{person}{Gregory Zelinsky}, \bibinfo{person}{Dimitris Samaras}, {and} \bibinfo{person}{Minh Hoai}.} \bibinfo{year}{2020}\natexlab{}.
\newblock \showarticletitle{Predicting Goal-Directed Human Attention Using Inverse Reinforcement Learning}. In \bibinfo{booktitle}{\emph{Proceedings of the IEEE/CVF Conference on Computer Vision and Pattern Recognition (CVPR)}}.
\newblock


\end{thebibliography}

%\input{source/07b2-shortened-ResultFigures}
%\input{source/07a2-shortened-ComparisonFigures}
%\input{source/07a3-ComparisonFiguresChen}
%\input{source/07c2-shortened_FewShots}
%\input{source/07g-IndividualResults}

% \input{source/07b-ResultFigures}
% \let\cleardoublepage\clearpage
% \vfill\eject %% <-- This fixes the blank page issue

%\balance

\end{document}